\documentclass[10pt,reqno,final]{article}

\usepackage{amsmath,amsfonts,amssymb,amsthm,version}
\usepackage{mathrsfs,fancybox,pifont}
\usepackage{graphicx}
\usepackage{url,hyperref}
\usepackage[notcite,notref]{showkeys}
\usepackage{color}
\usepackage{subfigure,multirow}
\usepackage{epstopdf}
\usepackage{cases}
\usepackage{mathtools}
\usepackage{algorithm,algorithmic}
\usepackage{authblk}
\usepackage{lipsum}
\usepackage{booktabs}
\usepackage{rotating}

\allowdisplaybreaks

\numberwithin{figure}{section}
\numberwithin{table}{section}

\theoremstyle{plain}
\newtheorem{theorem}{Theorem}[]
\newtheorem{lemma}{Lemma}[]
\newtheorem{corollary}{Corollary}[section]

\theoremstyle{definition}
\newtheorem{definition}{Definition}[section]

\theoremstyle{remark}
\newtheorem{remark}{Remark}[section]

\title{Adaptive incomplete multi-view learning via tensor graph completion}
\author[1]{Heng Zhang}
\author[1]{Xiaohong Chen\thanks{lyandcxh@nuaa.edu.cn}}
\affil[1]{College of Mathematics, Nanjing University of Aeronautics and Astronautics}

\date{\today}

\begin{document}
\maketitle

\begin{abstract}
	With the advancement of the data acquisition techniques, multi-view learning has become a hot topic. Some multi-view learning methods assume that the multi-view data is complete, which means that all instances are present, but this too ideal. Certain tensor-based methods for handing incomplete multi-view data have emerged and have achieved better result. However, there are still some problems, such as use of traditional tensor norm which makes the computation high and is not able to handle out-of-sample. To solve these two problems, we proposed a new incomplete multi-view learning method. A new tensor norm is defined to implement graph tensor data recover. The recovered graphs are then regularized to a consistent low-dimensional representation of the samples. In addition, adaptive weights are equipped to each view to adjust the importance of different views.  Compared with the existing methods, our method nor only explores the consistency among views, but also obtains the low-dimensional representation of the new samples by using the learned projection matrix. An efficient algorithm based on inexact augmented Lagrange multiplier (ALM) method are designed to solve the model and convergence is proved. Experimental results on four datasets show the effectiveness of our method.


\end{abstract}

\section{Introduction}

Multi-view data can provide more information to complete learning tasks compared with the single-view data in many realistic scenarios. Multi-view learning has become an active topic with the large amount of multi-view data being acquired. For different types of learning tasks and data forms, many multi-view learning methods have been proposed, which can be roughly summarized as multi-view transfer learning, multi-view dimension reduction, multi-view clustering, multi-view discriminant analysis, multi-view semi-supervised learning, and multi-view multi-task learning. For a comprehensive review of multi-view learning, please refer to \cite{MLORGNC}. Most previous methods work well under the assumption that all views of data are complete, i.e., all view representations of each sample are present. However, this assumption usually does not hold due to some instances being missing in practice. For example, in Alzheimer’s disease diagnosis \cite{IMRLADD}, the results of all examinations are not available to the subject for various reasons, which leads to incomplete multi-view data. In turn, incomplete multi-view learning has received attention of researchers.

Some incomplete multi-view learning methods have been developed, and the specific ones will be summarized in section \ref{relatedwork}. The graph-based incomplete multi-view learning method has attracted the attention of many researchers due to the fact that it is a powerful tool for analyzing the relationship among things \cite{DDGCGL}. Incomplete multi-view spectral clustering with adaptive graph learning (IMSC\_AGL) \cite{wenjieIMSCAG} uses graphs constructed from the low-rank subspace of each view to constrain the consensus low-dimensional representation of each sample. Generalized in complete multi-view clustering with flexible locality structure diffusion (GIMC\_FLSD) \cite{wenjieGIMCFLSD} introduces graph learning into matrix factorization to obtain unified representation, where the representation preserves the graph information. Incomplete multi-view non-negative representation learning with multiple graphs (IMNRL) \cite{IMNRLMG} performs matrix factorization on multiple incomplete graphs to obtain consensus non-negative representation with graph constraint. Adaptive graph completion based incomplete multi-view clustering (AGC\_IMC) \cite{wenjieAGCIMC} executes consensus representation learning using the self-representation of the graph of each view as regularization on incomplete multi-view data, and further gains the consensus low-dimensional representation. These methods all introduce graph learning to make fuller use of local geometric information among instances, but they only considered the similarity of un-missing instances of intra-view and ignored the similarity of inter-view, which made them unable to explore the complementary information embedded in multiple views well. 

In order to explore the high-order correlation information among views, some tensor-based methods are proposed to deal with incomplete multi-view data. Liu et al. \cite{IMSCLT} used subspace representations with low-rank tensor constraints to explore both the view-specific and cross-view relations among samples and capture the high-order correlations of multiple views simultaneously. Incomplete multi-view tensor spectral clustering with missing view inferring (IMVTSC-MVI) \cite{UTFIMCM} integrates feature space based on missing view inference and manifold space based on similarity graph learning into a unified framework, and introduces low-rank constraint to explore hidden missing view information and to capture high-order correlation information among views. It is worth noting that IMVTSC-MVI treats each view equally, that is, each view has the same weight. Xia et al. \cite{TCBIMC} obtained a complete graph for each view by taking into account the similarity of graphs among views and using the tensor completion tool, which makes full use of complementary information and spectral structure among views. Although the above tensor-based methods have achieved good performance, there are still some problems. Most of the existing work is to impose a low-rank constraint on the rotated graph tensor, which greatly increased the computational cost. This is because performing SVD on each transverse slice of the rotated tensor of $n\times m\times n$ is usually very time-consuming, where $n$ and $m$ are the number of samples and views respectively. To this end, a tensor nuclear norm based on semi-orthogonal matrix is redefined to reduce computation in this paper.

Specifically, our method is named adaptive incomplete multi-view learning via tensor graph completion (AIML\_TGC ), which integrates tensor graph completion, adaptive weight, and consistent subspace learning into a unified framework. Different above work, we redefine a tensor nuclear norm based on a semi-orthogonal matrix in order to reduce the computation of the solution and provide a shrinkage operator to obtain the analytical solution of the subproblem. AIML\_TGC has the unique advantage of subspace, that is, it can process out-of-samples to obtain its low-dimensional representation. To sum up, the contributions of this paper are summarized as follows:
\begin{enumerate}
	\item An incomplete multi-view learning method based on a tensor completion technique is proposed. Different from the existing multi-view learning method based on tensor norm, we redefine the tensor nuclear norm based on semi-orthogonal to reduce computational cost and assigned weights to each view adaptively.
	\item A shrinkage operator is provided to obtain a analytical solution to the subproblem with redefined nuclear norm. The solution method based on the inexact augmented Lagrange multiplier (ALM) is proposed, and its convergence is proved under mild conditions.
	\item Extensive experiments show that this method is effective in clustering tasks on four datasets.
	
\end{enumerate}
The structure of this paper is outlined as follows: Section \ref{sec2} introduces related work briefly. Section \ref{sec:main} proposes our incomplete multi-view learning method (AIML\_TGC), and gives the solving process in detail. Section \ref{sec:exper} verifies the effectiveness of the proposed method through a series of numerical experiments. Section \ref{sec:conc} summarizes the whole paper.

\section{Preliminaries and Related Work}
\label{sec2}
In this section, we summarize some of the notations and basic definitions that will be used in this paper, as well as relevant literature on processing incomplete multi-view data in recent years.

\subsection{Related Work}\label{relatedwork}

A large number of related work has evolved rapidly in the past few year. In this section, incomplete multi-view learning approaches and the use of tensor in multi-view learning will be briefly summarized.

We classify incomplete multi-view learning methods into five categories, i.e. subspace learning based methods, non-negative matrix factorization based methods, graph learning methods, deep neural network based methods, and probabilistic perspective based method. Each category is specified below. 
\begin{enumerate}
	\item Subspace learning based methods \cite{SemiLRKCCA}\cite{S2GCA}\cite{NeCA}\cite{PRMNeCA}\cite{SRRS}\cite{NMVL-IV}: These methods project all existing views into a common low-dimensional subspace and align all views in this subspace to obtain a consistent low-dimension representation. For example, Yang et al. \cite{SRRS} used sparse low-rank subspace representation to jointly measure relationships intraview relations and interview relations, and then proposed an incomplete multi-view dimension reduction method. Chen et al. \cite{S2GCA} proposed semi-paired and semi-supervised generalized correlation analysis (S$^2$GCA) based on canonical correlation analysis (CCA) , which not only processes incomplete data, but also is applicable to semi-supervised scenario.
	\item Non-negative matrix factorization (NMF) based methods \cite{PVC}\cite{MVIIV}\cite{wenjieIMCGRMF}\cite{DAIMC}\cite{OPIMVC}\cite{wenjieGIMCFLSD}\cite{NCLAIMC}\cite{IMNRLMG}: These methods aim to decompose the data matrix of each view into the product of the corresponding view’s basis matrix and common representation matrix, and then proceed with the next step based on the representation matrix. For example, Li et al. \cite{PVC} proposed NMF based partial multi-view clustering to learn a common representation for complete samples and private latent representation with the same basis matrix. Multi-view learning with incomplete views (MVL-IV) \cite{MVIIV} factorizes the incomplete data matrix of each view as the product of a common representation matrix and corresponding view’s basis matrix with low-rank property.
	\item Graph learning based methods \cite{MKKMIK}\cite{wenjieAGCIMC}\cite{wenjieIMSCAG}\cite{wenjieCGIMSC}\cite{MVSCIG}\cite{LFIMVC}\cite{APGLFIMC}\cite{APGLFIMVC}: These methods use the consensus among views to construct graph matrix (or kernel matrix) so that the geometry structure can be preserved, which mainly includes spectral clustering and multi-kernel learning. For example, adaptive graph completion based incomplete multi-view clustering (AGC\_IMC) \cite{wenjieAGCIMC} obtains common graph by introducing the self-representation regularization term of the graph matrix, and further gains the consensus low-dimensional representation of the samples. Incomplete multiple kernel k-means with mutual kernel completion (MKKM-IK-MKC) \cite{MKKMIK} adaptively imputes incomplete kernel matrices, and combines imputation and clustering to achieve higher clustering performance.
	\item Deep neural network based methods \cite{IMVSCAISMDFF}\cite{IMVCDSM}\cite{CMPNET}\cite{PMVCCGAN}\cite{AIMSCN}: Theses methods make use of neural network’s powerful learning ability to obtain deeper and complete feature representation. For example, Zhao et al. \cite{IMVCDSM} developed a novel incomplete multi-view clustering method, which projects all multi-view data to a complete and unified representation with the constrain of intrinsic geometric structure. Wang et al. \cite{PMVCCGAN} and Xu et al. \cite{AIMSCN} proposed incomplete multi-view clustering methods based on generative adversarial network (GAN) \cite{GAN}, aiming at two-view and multi-view data respectively, i.e., they use existing views to generate missing views. Lin et al. \cite{COMPLETER} introduced the ideal of contrast prediction from self-supervised learning in to incomplete multi-view learning to explore the consensus between views by maximizing mutual information and minimizing conditional entropy.
	\item Probabilistic perspective based methods \cite{SPPCCA}\cite{GBCCAMM}\cite{SGPLVMIMC}\cite{PSCCA}: Theses methods usually use Baysian statistics, Gaussian models, Variational inference, and other tools to model multi-view data from a probabilistic perspective. For example, Li et al. \cite{SGPLVMIMC} used shared gaussian process latent variable model to model the incomplete multi-view data for clustering. Matsuura et al. proposed generalized bayesian canonical correlation analysis with missing modalities (GBCCA-M2) \cite{GBCCAMM} by combining Bayesian CCA \cite{VBACCA} and Semi-CCA \cite{SCCAESLCC} to process incomplete multi-view data with high dimension.
\end{enumerate}

There are two ways to use tensor in the field of multi-view learning. One ways is to look at multi-view data as higher order data and process them by some higher order data methods. The other is to use the method of processing tensor to capture the higher information among views. A representative approach in the first way is tensor canonical correlation analysis (TCCA) \cite{TCCA}. TCCA defines a covariance tensor to handle multi-view data with arbitrary number of views by extending the mutual covariance matrix in canonical correlation analysis (CCA). Deep TCCA (DTCCA) \cite{DEEPTCCA} is further proposed by combining it with deep learning. Unlike kernel TCCA, DTCCA can not only handle multi-view data with arbitrary number of views in a nonlinear manner, but also does not need to maintain the train data for computing representations of any given data, which means that a large amount of data can be handled. Cheng et al. \cite{TBDRLMVC} viewed the multi-view data as a 3-order tensor and constructed the self-representation tensor by reconstructing the coefficients. Further the Tucker decomposition is used to obtain a low-dimensional representation of samples. The second way usually assumes that the graphs or representation weight matrices of different view are similar, and thus the tensor composed of them is assumed to be of low rank. Low-rank Tensor constrained Multiview Subspace Clustering (LT-MSC) \cite{LRTCMSC} regards the subspace representation matrices of different views as a tensor to capture the higher-order consensus information among views. And this tensor is equipped with low-rank constrains to reduce the redundancy of the learned subspaces and improve the clustering performance. Wu et al. \cite{ETLMVSC} combined the ideal of robust principle component analysis using tensor singular value decomposition based tensor nuclear norm to preserve the low-rankness of a tensor constructed based on a multi-view transition probability matrices of the Markov chain. Kernel $k$-means coupled graph tensor (KCGT) \cite{MKCKKCGTL} uses kernel matrices to construct graph with symmetry and properties of block diagonal. Further these graphs are stacked into a low-rank graph tensor for capturing the high order affinity of all these graphs.


\subsection{Preliminaries}
 
Tensor and matrices  are denoted by uppercase  calligraphic letters and italic capital letters respectively, e.g. $\mathcal{X}\in\mathbb{R}^{n_1\times n_2\times n_3}$ and $X\in \mathbb{R}^{n_1\times n_2}$. We denote $\mathcal{X}_{\left(i,j,k\right)}$ and $X_{\left(i,j\right)}$ are $i,j,k$-th element of $\mathcal{X}$ and $i,j$-th element of $X$ respectively. Some basic definitions of tensor are shown below.

\begin{table}[htbp]
	\small
	\centering
	\caption{Some notations and descriptions used in this article.}
	\begin{tabular}{cp{12cm}}
		\toprule
		Notations & Descriptions \\ 
		\midrule
		$\mathcal{X}_{(i,j,k)}$& The $i,j,k$-th element of the tensor $\mathcal{X}$.\\
		$X_{(i,j)}$ & The $i,j$-th element of the matrix $X$. \\ 
		$\mathcal{X}_{(:,:,i)}$ or $X^{(i)}$ & The $i$-th frontal slice of the tensor $\mathcal{X}.$\\
		$X_{(:,i)}$ & The $i$-th column of the matrix $X$.\\
		$X^\top$ & The transpose of the matrix $X$.\\
		$\left\|X\right\|_*$ & The nuclear norm of the matrix $X$.\\
		$\left\|X\right\|_F$ & The Frobenius norm of the matrix $X$.\\
		$\left\|\mathcal{X}\right\|_F$ & The Frobenius norm of the tensor $\mathcal{X}$, i.e. $\sqrt{\sum_{i,j,k}\mathcal{X}_{\left(1,j,k\right)}^2}$.\\
		$\sigma_i\left(X\right)$ & The $i$-th smallest singular value of the matrix $X$.\\
		$\left\langle\mathcal{X},\mathcal{Y}\right\rangle$& The inner product of the tensor $\mathcal{X}$ and $\mathcal{Y}$, i.e. $\sum_{i,j,k}\mathcal{X}_{\left(i,j,k\right)}\mathcal{Y}_{\left(i,j,k\right)}$.\\
		$\nabla f\left(x\right)$& The gradient of the function $f \left(x\right)$.\\
		$x_i$ & The $i$-th element of the vector $x$.\\		
		\bottomrule 
	\end{tabular}
	\label{tab:note}
\end{table}

\begin{definition}[\emph{unfold} and \emph{fold}]
	For a 3-order tensor  $\mathcal{X} \in\mathbb{R}^{n_1\times n_2\times n_3}$, we have
	\begin{equation}
		\emph{unfold}\left(\mathcal{X}\right) =\left[\begin{matrix}
			X^{(1)}\\ X^{(2)}\\ \vdots\\ X^{(n_3)}
		\end{matrix}\right]=\left[X^{(1)}; X^{(2)}; \cdots; X^{(n_3)}\right]\in\mathbb{R}^{n_1n_3\times n_2},
	\end{equation}
	\begin{equation}
		\emph{fold}\left(\emph{unfold}\left(\mathcal{X}\right)\right) = \mathcal{X}.
	\end{equation}
\end{definition} 

\begin{definition}[transformed tensor of $\mathcal{X}$]
	For a 3-order tensor $\mathcal{X}\in \mathbb{R}^{n_1 \times n_2 \times n_3}$, given an matrix $\Phi\in \mathbb{R}^{n_3\times n_3}$, the transformed tensor of the tensor $\mathcal{X}$ is defined as
	\begin{equation}
		\mathcal{X}_\Phi = \emph{fold}\left(\left[\begin{matrix}
			\mathcal{X}_{(:,1,:)}\\ \mathcal{X}_{(:,2,:)}\\\vdots\\\mathcal{X}_{(:,n_1,:)}
		\end{matrix}\right]\Phi\right)\in \mathbb{R}^{n_1\times n_2\times n_3}	.
	\end{equation}
	
\end{definition}

\begin{definition}[$\Phi$-Transformed Tensor Nuclear Norm, $\Phi$-TTNN]
	For a 3-order tensor $\mathcal{X}\in\mathbb{R}^{n_1\times n_2\times n_3}$, given an orthogonal matrix $\Phi\in \mathbb{R}^{n_3\times n_3}$, the $\Phi$-TTNN of the tensor $\mathcal{X}$ is defined as
	\begin{equation}
		\left\|\mathcal{X}\right\|_{\Phi-TTNN} = \sum\limits_{i=1}^{n_3}\left\|{X_\Phi}^{(i)}\right\|_*.
	\end{equation}

\end{definition}

\section{Proposed Method}
\label{sec:main}
\subsection{The proposed model}

CCA is classical and effective method for learning multiple view, where the goal is to find two projection vector for two views in order to maximize the correlation between the two projected views. Note that CCA can handle data from only two views, for this reason generalized CCA (GCCA) \cite{GCCA} is proposed to handle data from multiple views. Specifically, given a multi-view dataset $\left\{X_i\in\mathbb{R}^{d_i\times n}\right\}_{i=1}^m$, the model of GCCA is as follows,
\begin{equation}
	\label{GCCA}
	\begin{split}
		\underset{A,\left\{W_i\right\}_{i=1}^{m}}{\mathop{\min}}&\quad\sum\limits_{i=1}^{m}{\left\|A-W_i^\top X_i\right\|_F^2}\\
		\text{s.t.}&\quad AA^\top=I
	\end{split}.
\end{equation}
Symbol $A$ and $\left\{W_i\right\}_{i=1}^m$ in Eq.(\ref{GCCA}) are the consistent low-dimensional representation matrix of the samples and the projection matrix of each view, respectively. The geometric structure among samples is a valid priori information for the subsequent tasks. As mentioned in the introduction, graph can be exploring the relationship between things. For this reason, graph multi-view CCA (GMCCA) \cite{GMCCA} is proposed to obtain consistent low-dimensional representation with structure preservation, and its objective function is 

\begin{equation}
	\label{GMCCA}
	\begin{split}
		\underset{A,\left\{W_i\right\}_{i=1}^{m}}{\mathop{\min}}&\quad\sum\limits_{i=1}^{m}{\left\|A-W_i^\top X_i\right\|_F^2}+\lambda\operatorname{tr}\left(AL_GA^\top\right)\\
		\text{s.t.}&\quad AA^\top=I
	\end{split},
\end{equation}
where $L_G$ is the preconstructed graph Laplacian matrix and $\lambda$ is the regularization parameter. 

A model capable of handing incomplete multi-view data is designed based on Eq.(\ref{GMCCA}) and combined with tensor recover in this paper. For a given incomplete multi-view dataset with $m$ views and $n$ samples $\left\{X_i\in\mathbb{R}^{d_i\times n}\right\}_{i=1}^m$, whose missing instance filled by $\textbf{0}$ (i.e. if $j$-th instance of $i$-th view is missing, then $X_{i\left(:,j\right)}=\textbf{0}$) and the missing information of each view are recorded in a diagonal matrix $\left\{P_i\in\mathbb{R}^{n\times n}\right\}_{i=1}^m$. $P_i$ is defined as follows,
\begin{equation}
	\nonumber
	\label{equ11}
	P_{i(j,j)}=\left\{\begin{split}
		&1 \quad
			\text{if } j\text{-th instance of the }i\text{-th view is not missing}\\
		&0 \quad\text{otherwise} 
	\end{split} \right..
\end{equation}
Symbol $d_i$ is feature dimension of the $i$-th view. Denote $n_i$ as the number of available instance of the $i$-th. Based on Eq.(\ref{GMCCA}), a naive model for handing incomplete multi-view data is designed as follows,
\begin{equation}
	\label{naive}
	\begin{split}
		\underset{A,\left\{W_i\right\}_{i=1}^{m}}{\mathop{\min}}&\quad\sum\limits_{i=1}^{m}\left(\left\|\left(A-W_i^\top X_i\right)P_i\right\|_F^2+\lambda\operatorname{tr}\left(AL_{G_i}A^\top\right)\right)\\
		\text{s.t.}&\quad AA^\top=I
	\end{split}.
\end{equation}
where $L_{G_i}$ is the Laplacian matrix of preconstructed graph $G_i$ of the $i$-th view. It is worth noting that model (\ref{naive}) treats each view equally, which is usually unrealistic. For this reason, inspired by Nie et al. \cite{SWMCMG}, weight $\bar{\delta}_i$ are introduced to distinguish the importance of different views as follows,
\begin{equation}
	\label{weight}
	\bar{\delta}_i=\frac{1}{\sqrt{\left\|\left(A-W_i^\top X_i\right)P_i\right\|_F^2+\lambda\operatorname{tr}\left(AL_{G_i}A^\top\right)}}.
\end{equation}
Intuitively, if $i$-th view is good, then $\left\|\left(A-W_i^\top X_i\right)P_i\right\|_F^2$ and $\operatorname{tr}\left(AL_{G_i}A^\top\right)$ should be small, and thus the weight $\bar{\delta}_i$ for $i$-th view is large accordingly Eq.(\ref{weight}). However, it is worth noting that if a view has a high missing rate, it will also make the weight that view larger. To avoid this phenomenon we improve the weight to $\delta_i=n_i\bar{\delta}_i$.

\begin{figure}[htbp]
	\begin{center}
		\includegraphics[width=6.5in]{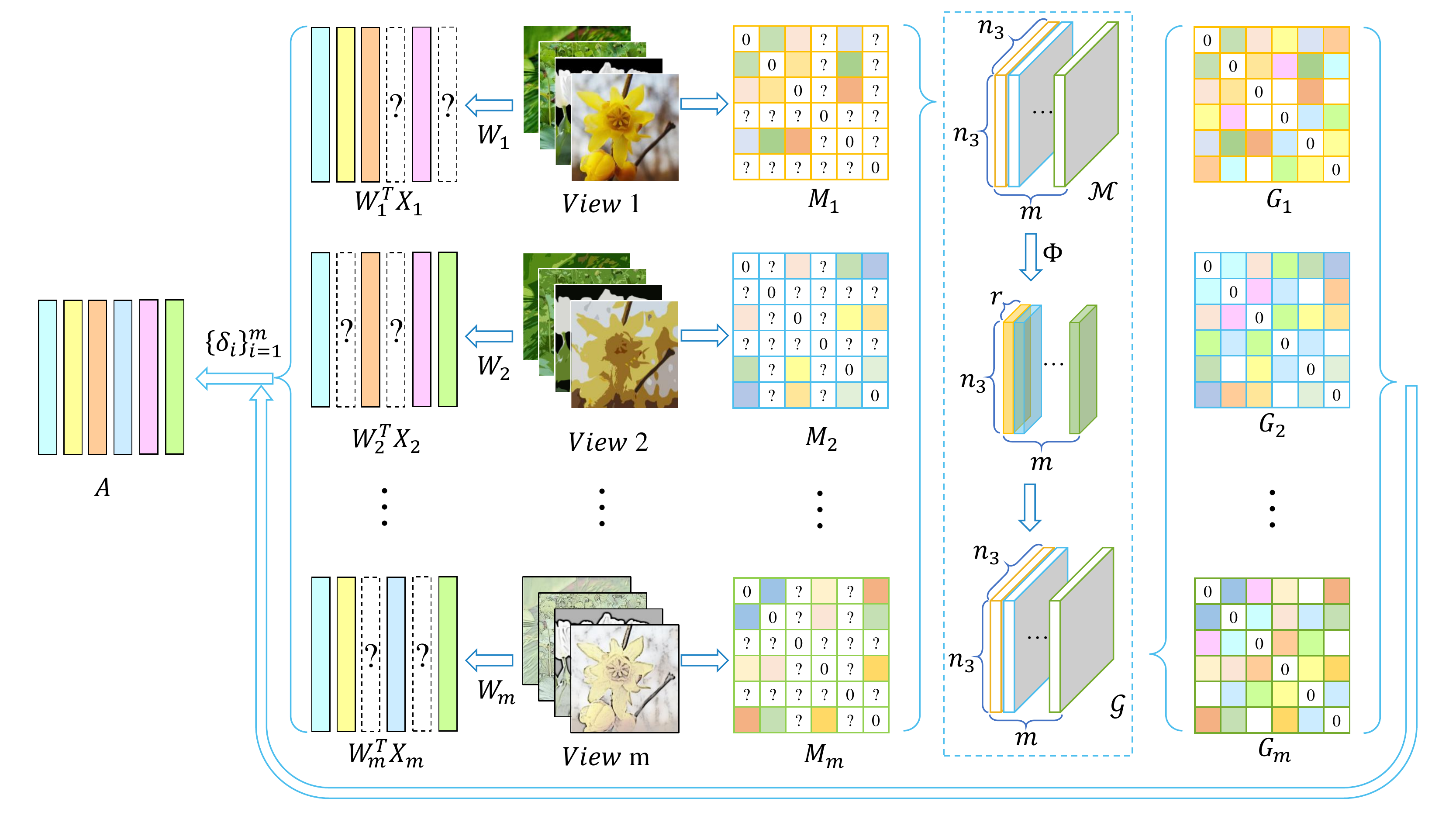}
		\caption{The framework of the proposed method.}
		\label{fig1}
	\end{center}
\end{figure}

Unfortunately, it is not possible to directly construct a complete graph for each view due to the incompleteness of incomplete multi-view data. Therefore, $G_i$ needs to be completed. Consistency among views is an important property of multi-view data. The graphs among different views should be highly similar in response to consistency among views. Drawing inspiration from the filed of tensor recovery, the objective function for graph recovery is as follows,
\begin{equation}
	\label{graphrecover}
	\begin{split}
		\underset{\mathcal{G}}{\min}&\quad\left\|\mathcal{G}\right\|_{\Phi,w,S_p}^p+\frac{\gamma}{2}\left\|P_{w}\left(\mathcal{G}\right)-P_{w}\left(\mathcal{M}\right)\right\|_F^2\\
		\text{s.t.}&\quad G_i\geq 0,\quad G_i\textbf{1}=\textbf{1}
	\end{split}
\end{equation}
where $\mathcal{G}\in\mathbb{R}^{n\times m\times n}$ and $\mathcal{G}_{\left(:,i,:\right)}=G_i$. $\mathcal{M}\in\mathbb{R}^{n\times m\times n}$ is the preconstructed incomplete graph tensor with missing positions filled by 0, where $\mathcal{M}_{\left(:,i,:\right)}=M_i$ is the preconstructed graph of the $i$-th view. $P_w\left(\cdot\right)$ is the orthogonal projection onto the liner subspace of tensors supported on $w=\left\{\left(i,j,k\right)|\mathcal{M}_{\left(i,j,k\right)} \text{ is not missing}\right\}$: $P_w\left(\mathcal{M}\right)_{\left(i,j,k\right)}=\mathcal{M}_{\left(i,j,k\right)}$ if $\left(i,j,k\right)\in w$ and $P_w\left(\mathcal{M}\right)_{\left(i,j,k\right)}=0$ otherwise. $\left\|\cdot\right\|_{\Phi,w,S_p}$ is redefined tensor norm as follows,

\begin{definition}[$\Phi$-Transformed weight Schatten-$p$ Tensor Nuclear Norm]
	Given tensor $\mathcal{X}\in\mathbb{R}^{n_1\times n_2\times n_3}$, $h=\operatorname{min}\left(n_1,n_3\right)$, weight vector $w\in\mathbb{R}^h$ with $w_{1} \geq w_{2} \geq \cdots \geq w_{l}\geq0$, $0<p\leq 1$, and transformed matrix $\Phi\in\mathbb{R}^{n_3\times r}$, then
	\begin{equation}
		\left\|\mathcal{X}\right\|_{\Phi,w,S_p}=\left(\sum\limits_{i=1}^{n_3}\left\|X_{\Phi}^{\left(i\right)}\right\|_{w,S_p}^p\right)^{\frac{1}{p}}=\left(\sum\limits_{i=1}^{n_3}\sum\limits_{j=1}^hw_j\sigma_j\left(X_{\Phi}^{\left(i\right)}\right)^p\right)^{\frac{1}{p}}.
	\end{equation}
\end{definition}
\begin{remark}
	Note that in some previous literature, the transform matrix is usually an orthogonal square matrix, i.e., $\Phi^\top\Phi = I$ and $r = n_3$. Our empirical study shows that this is not necessary. Therefore, a semi-orthogonal matrix is used as the transform matrix in this paper, i.e., $\Phi^\top\Phi = I$ and $r<n_3$. This will reduce the computational effort, i.e., $n_3$ singular value decomposition were required, but now only $r$ are needed. The experimental study in section \ref{norm_nn} shows that $r=n_3$ may not be optimal.
\end{remark}
Combining the graph recovery model (\ref{graphrecover}) with Eq.(\ref{naive}) and introducing weights $\left\{\delta_i\right\}_{i=1}^m$ gives our proposed method as follow,
\begin{equation}
	\label{our}
	\begin{split}
		\underset{A,\left\{W_i\right\}_{i=1}^m\mathcal{G}}{\min}&\quad \sum\limits_{i=1}^mn_i\sqrt{\left\|\left(A-{W_i}^\top X_i\right)P_i\right\|_F^2+\lambda \operatorname{tr}\left(AL_{G_i}A^\top\right)}+\mu\left\|\mathcal{G}\right\|_{\Phi,w,S_p}^p+\frac{\gamma}{2}\left\|P_{w}\left(\mathcal{G}\right)-P_{w}\left(\mathcal{M}\right)\right\|_F^2\\
		\text{s.t.}&\quad AA^\top=I,\quad G_i\geq 0,\quad G_i\textbf{1}=\textbf{1}
	\end{split},
\end{equation}
where $\mu>0$ is the regularization parameter. The pipeline of the proposed method is plotted in Fig.\ref{fig1}.
\begin{remark}
	Lemma \ref{lemma1} shows that the number of zero eigenvalues of the Laplacian matrix $L_G$ should be constrained to $c$ if a graph $G$ with $c$ number of connected components is obtained.  That is, $\sum_{i=1}^{c}\sigma_i\left(L_G\right)$ should equal zeros. According to Ky Fan's theory \cite{KYFANS}, the following equation holds, 
	\begin{equation}
		\sum\limits_{i=1}^{c}\sigma_i\left(L_G\right) = \underset{FF^\top=I,F\in\mathbb{R}^{c\times n}}{\min} \operatorname{tr}\left(FL_GF^\top\right).
	\end{equation}
	This is consistent with the second term in Eq.(\ref{our}), which implies that our method can be learned a consistent low-dimensional representation that is more conductive to clustering. It is worth noting that the presence of the first term of Eq.(\ref{our}) allows our method to handle out-of-sample.
\end{remark}
\begin{lemma}\cite{ELRCG}
	\label{lemma1}
	The number of zero eigenvalues of Laplacian matrix $L_G$ is equal to the number of connected components of the graph $G$. 
\end{lemma}
\subsection{Solution}
To efficiently solve our challenging problem (\ref{our}), we need to introduce the auxiliary variable $\mathcal{Y}$ to split the interdependent term such that they can be solved independently. Thus we can reformulate problem (\ref{our}) into the following equivalent form,
\begin{equation}
	\label{our1}
	\begin{split}
		\underset{A,\left\{W_i\right\}_{i=1}^m\mathcal{G}}{\min}&\quad \sum\limits_{i=1}^m\left(\delta_i\left\|\left(A-{W_i}^\top X_i\right)P_i\right\|_F^2+\lambda \delta_i\operatorname{tr}\left(AL_{G_i}A^\top\right)\right)+\mu\left\|\mathcal{Y}\right\|_{\Phi,w,S_p}^p+\frac{\gamma}{2}\left\|P_{w}\left(\mathcal{G}\right)-P_{w}\left(\mathcal{M}\right)\right\|_F^2\\
		\operatorname{s.t.}&\quad AA^\top=I,\quad G_i\geq 0,\quad G_i\textbf{1}=\textbf{1},\quad \mathcal{G}=\mathcal{Y}, \quad \delta_i=\frac{n_i}{\sqrt{\left\|\left(A-{W_i}^\top X_i\right)P_i\right\|_F^2+\lambda \text{tr}\left(AL_{G_i}A^\top\right)}}
	\end{split}
\end{equation}

Inspired by recent progress on alternating direction methods, we proposed efficient algorithm based on inexact augmented Lagrange multiplier (ALM) method to solve the problem (\ref{our1}), whose augmented Lagrangian function is given by
\begin{equation}
	\label{largfun}
	\begin{split}
		L_\rho\left(A,\left\{W_i\right\}_{i=1}^m,\mathcal{G},\mathcal{Y},\mathcal{C}\right)&=\sum\limits_{i=1}^m\left[\delta_i\left\|\left(A-{W_i}^\top X_i\right)P_i\right\|_F^2+\lambda \delta_i\operatorname{tr}\left(AL_{G_i}A^\top\right)\right]+\mu\left\|\mathcal{Y}\right\|_{\Phi,w,S_p}^p\\
		&+\frac{\gamma}{2}\left\|P_{w}\left(\mathcal{G}\right)-P_{w}\left(\mathcal{M}\right)\right\|_F^2+\left\langle\mathcal{C}, \mathcal{G}-\mathcal{Y}\right\rangle+\frac{\rho}{2}\left\|\mathcal{G}-\mathcal{Y}\right\|_F^2\\
	\end{split}
\end{equation}
where $\rho>0$ is the penalty parameter, $\mathcal{C}\in\mathbb{R}^{n\times m\times n}$ is Lagrange multipliers. This section uses superscripts to indicate the number of iterations, i.e. $A^k$ or $A^{k+1}$, etc.

\subsubsection{Updating $A^{k+1}$ and $\left\{W_i^k+1\right\}_{i+1}^m$}
To update $A^{k+1}$ and $\left\{W_i^{k+1}\right\}_{i=1}^m$, we consider the following optimization problems,
\begin{equation}
	\label{updataAW}
	\begin{split}
		A^{k+1}, \left\{W_i^{k+1}\right\}_{i=1}^m=\underset{A,\left\{W_i\right\}_{i=1}^m}{\arg\min}&\quad\sum\limits_{i=1}^m\left(\delta^k_i\left\|\left(A-{W_i}^\top X_i\right)P_i\right\|_F^2+\lambda \delta^k_i\operatorname{tr}\left(AL_{G^k_i}A^\top\right)\right)\\
		\operatorname{s.t.}&\quad AA^\top=I
	\end{split}.
\end{equation}
After simple algebraic operations, problem (\ref{updataAW}) can be transformed into the following eigenvalue problem,
\begin{equation}
	\label{updateA}
	\left(\sum\limits_{i=1}^m\delta_i^k\left(P_i-P_iX_i^\top\left(X_iP_iX_i^\top\right)^{-1}X_iP_i+\lambda L_{G_i^k}\right) \right){A}^\top={A}^\top\Sigma,
\end{equation}
where $\Sigma$ is a diagonal matrix of eigenvalues. Then $A^\top$ can be obtained from the eigenvector corresponding to the first $d$ largest eigenvalues, where $d$ is the dimension of consistent low-dimensional representation (i.e. $A\in\mathbb{R}^{d\times n}$). With the $A$, we can obtain $W_i^{k+1}=\left(X_iP_iX_i^\top\right)^{-1}X_iP_i{A^{k+1}}^\top$ for $i=1,2,\dots,m$. Once $A^{k+1}$ and $\left\{W_i^{k+1}\right\}_{i=1}^m$ are obtained, the $\left\{\delta_i\right\}_{i=1}^m$ can be updated according to the definition to obtain $\left\{\delta_i^{k+1}\right\}_{i=1}^m$.

\subsubsection{Updating $\mathcal{G}^{k+1}$}\label{sec:updateG}
To solve $\mathcal{G}^{k+1}$, we fix the other variable and solve the following optimization problem, 
\begin{equation}
	\label{updateG}
	\begin{split}
		\mathcal{G}^{k+1}=\underset{\mathcal{G}}{\arg\min}&\quad\lambda\sum\limits_{i=1}^m\delta_i^{k+1}\operatorname{tr}\left(A^{k+1}L_{G_i}{A^{k+1}}^\top\right)+\frac{\gamma}{2}\left\|P_w\left(\mathcal{G}\right)-P_w\left(\mathcal{M}\right)\right\|_F^2+\frac{\rho^k}{2}\left\|\mathcal{G}-\mathcal{Y}^k+\frac{\mathcal{C}^k}{\rho^k}\right\|_F^2\\
		\operatorname{s.t.}& \quad G_i\geq 0,\quad G_i\textbf{1}=\textbf{1}
	\end{split}
\end{equation}
Problem (\ref{updateG}) is a least squares problem with constrains. Notice that each tube of the tensor of problem (\ref{updateG}) is independent, so it can be converted into $n\times m$ independent subproblems. Without loss of generality, we let $g\in\mathbb{R}^{n_3}$, $s\in\mathbb{R}^{n_3}$, $y\in\mathbb{R}^{n_3}$, $\omega\in\mathbb{R}^{n_3}$, $c\in\mathbb{R}^{n_3}$, and $t\in\mathbb{R}^{n_3}$ be some tubes of tensor $\mathcal{G}$, $\mathcal{M}$, $\mathcal{Y}$, $w$, $\mathcal{C}$, and $\mathcal{T}$, respectively, where  $\mathcal{T}^k_{\left(i,j,l\right)}=\frac{1}{2}\lambda\delta_l^{k+1}\left\|A^{k+1}_{\left(:,i\right)}-A^{k+1}_{\left(:,j\right)}\right\|_F^2$. Thus the following optimization problem can be obtained,
\begin{equation}
	\label{updateg}
	\begin{split}
		g^{k+1}=\underset{g}{\arg\min}&\quad \frac{1}{\rho^k}{t^k}^\top g+\frac{\gamma}{2\rho^k}\left\|P_\omega\left(g\right)-P_\omega\left(z\right)\right\|_F^2+\frac{1}{2}\left\|g-y^k+\frac{c^k}{\rho^k}\right\|_F^2\\
		\operatorname{s.t.}& \quad g\geq 0,\quad \textbf{1}^\top g=1
	\end{split}.
\end{equation}
The Lagrangian function of Eq.(\ref{updateg}) is 
\begin{equation}
	L\left(g, \alpha, \beta\right) = \frac{1}{\rho^k}{t^k}^\top g+\frac{\gamma}{2\rho^k}\left\|P_\omega\left(g\right)-P_\omega\left(z\right)\right\|_F^2+\frac{1}{2}\left\|g-y^k+\frac{c^k_1}{\rho^k}\right\|_F^2 - \alpha\left(\textbf{1}^\top g-1\right)-\beta^\top g. 	
\end{equation}
where $\alpha\in\mathbb{R}$ and $\beta\in\mathbb{R}^{n_3}$ are multipliers. Taking the derivative w.r.t. $g$ and setting it to zero, the following equation holds, 
\begin{equation}\label{dge0}
	g-y^k+\frac{c^k_1}{\rho^k}+\frac{\gamma}{\rho^k}\left(P_\omega\left(g\right)-P_\omega\left(z\right)\right)+\frac{1}{\rho^k}t^k - \alpha\textbf{1}-\beta=\textbf{0}.
\end{equation}
The according to the KKT condition, i.e., $\textbf{1}^\top g=1$, we have 
\begin{equation}
	\label{aplha}
	\alpha=\frac{1}{n}\left(1-\textbf{1}^\top y^k+\textbf{1}^\top\frac{c^k_1}{\rho^k}+\frac{\gamma}{\rho^k}\textbf{1}^\top\left(P_\omega\left(g\right)-P_\omega\left(z\right)\right)+\frac{1}{\rho^k}\textbf{1}^\top t^k - \textbf{1}^\top\beta\right).
\end{equation}
Combining Eq.(\ref{dge0}) and Eq.(\ref{aplha}), we can obtain
\begin{equation}
	\begin{split}
		g+\frac{\gamma}{\rho^k}P_\omega\left(g\right)=&y^k-\frac{c^k_1}{\rho^k}+\frac{\gamma}{\rho^k}P_\omega\left(z\right)-\frac{1}{\rho^k}t^k - \frac{1}{n}\textbf{1}\textbf{1}^\top\left( y^k-\frac{c^k_1}{\rho^k}+\frac{\gamma}{\rho^k}+P_\omega\left(z\right)-\frac{1}{\rho^k}t^k\right)+\frac{1}{n}\textbf{1}\\ &+\frac{1}{n}\textbf{1}\textbf{1}^\top\left(\frac{\gamma}{\rho^k}P_\omega\left(g\right)-\beta\right)+\beta
	\end{split}
\end{equation}
Further according to the complementary relaxation in the KKT condition, we have
\begin{equation}
	g^{k+1}_i+\frac{\gamma}{\rho^k}P_\omega\left(g^{k+1}\right)_i=\left(u^k_i-\frac{1}{n}\textbf{1}^\top u^k+\frac{1}{n}+v^k\right)_+
\end{equation}
where $u^k=y^k-\frac{c^k_1}{\rho^k}+\frac{\gamma}{\rho^k}P_\omega\left(z\right)-\frac{1}{\rho^k}t^k$, $v^k=\frac{1}{n}\textbf{1}^\top\left(\frac{\gamma}{\rho^k}P_\omega\left(g^{k+1}\right)-\beta\right)$. So we can get
\begin{equation}
	\label{updategk}
	g^{k+1}_i=\left\{\begin{split}
		&\left(u^k_i-\frac{1}{n}\textbf{1}^\top u^k+\frac{1}{n}+v^k\right)_+,\qquad \qquad \text{if } i \in \omega,\\
		&\frac{\rho^k}{\rho^k+\gamma}\left(u^k_i-\frac{1}{n}\textbf{1}^\top u^k+\frac{1}{n}+v^k\right)_+, \quad \text{otherwise}.
	\end{split}\right.
\end{equation}
Since $\textbf{1}^\top g^{k+1}=1$, the value of $v^k$ can be found by solving the roots of the following function.
\begin{equation}
	\label{sov}
	f\left(v\right)=\sum\limits_{i=1}^ng_i^{k+1}-1
\end{equation}
Equation (\ref{sov}) can be solved by Newton's method. The first order derivative of $f\left(v\right)$ is 
\begin{equation}
	\label{gradig}
 	\nabla f\left(v\right)=\sum\limits_{i=1}^n\nabla g_i^{k+1},
\end{equation}
where 
\begin{equation}
	\nabla g_i^{k+1}=\left\{\begin{split}
		&1,\qquad\qquad \text{if } i \in \omega,\\
		&\frac{\rho^k}{\rho^k+\gamma},\quad \text{otherwise}.
	\end{split} \right.
\end{equation}
The solution procedure for $g$ is organized as in Algorithm \ref{alg:g}.
\begin{algorithm}
	\caption{Solving problem \ref{updateg}.}
	\label{alg:g}
	\begin{algorithmic}
		\STATE{\textbf{Input:} $\left\{X_i\right\}_{i=1}^m$, $\left\{P_i\right\}_{i=1}^m$, $\omega$, $\mathcal{M}$, $\mu$, $\lambda$.}
		\STATE{\textbf{Initialize:} $u=y^k-\frac{c^k_1}{\rho^k}+\frac{\gamma}{\rho^k}P_\omega\left(z\right)-\frac{1}{\rho^k}t^k$, $v^0=0$, $k=0$}
		\WHILE{$|f\left(v\right)|>10^{-10}$}
		\STATE{Evaluate the function $f\left(v^k\right)$ according to Eq.(\ref{sov}).}
		\STATE{Compute the gradient of $f\left(v\right)$ at point $v^k$ according to Eq.(\ref{gradig}).}
		\STATE{Update $v$ by $v^{k+1}=v^k-\frac{f\left(v^k\right)}{\nabla f\left(v^k\right)}$}
		\STATE{$k\leftarrow k+1$}
		\ENDWHILE
		\STATE{\textbf{Output:} $g$ via Eq.(\ref{updategk})}
	\end{algorithmic}
\end{algorithm}

\subsubsection{Updating $\mathcal{Y}^{k+1}$}\label{sec:updateY}
By keeping all other variable fixed, $\mathcal{Y}^{k+1}$ can be updated by solving the following optimization problem,
\begin{equation}
	\label{updateY}
	\underset{\mathcal{Y}}{\min}\quad\frac{\mu}{\rho^k}\left\|\mathcal{Y}\right\|_{\Phi,w,S_p}^p+\frac{1}{2}\left\|\mathcal{G}^{k+1}+\frac{\mathcal{C}^k}{\rho^k}-\mathcal{Y}\right\|_F^2.
\end{equation}

To solve it, we first introduce the Theorems \ref{thm1}, which is proved in the Appendix \ref{A1}.
\begin{theorem}\label{thm1}
	Suppose $\mathcal{A} \in \mathbb{R}^{n_{1} \times n_{2} \times n_{3}}$, $l=\min \left(n_{1}, n_{2}\right)$, $w_{1} \geq w_{2} \geq\cdots \geq w_{l}\geq0$. Given the model
	\begin{equation}
		\label{thm1equ}
		\underset{\mathcal{X}}{\min}\quad \tau\|\mathcal{X}\|_{\Phi, w, S_{p}}^{p}+ \frac{1}{2}\|\mathcal{X}-\mathcal{A}\|_{F}^{2},
	\end{equation}
	then the optimal solution to the model (\ref{thm1equ}) is
	\begin{equation}
		\mathcal{X}^{*}=\left(\operatorname{fold}\left(\operatorname{unfold}\left(\mathcal{S}_{\tau,w,p}(\mathcal{A}_\Phi)\right);\operatorname{unfold}\left(\mathcal{A}_{\Phi^c}\right)\right)\right)_{\overline{\Phi}^\top},
	\end{equation}
	where $\mathcal{S}_{\tau,w,p}\left(\mathcal{A}_\Phi\right)$ is a tensor, which the $i$-th frontal slice is $S_{\tau,w,p}\left(A_\Phi^{(i)}\right)$.
\end{theorem}
Therefore, the optimal solution of Eq.(\ref{updateY}) is 
\begin{equation}
	\label{updateY2}
	\mathcal{Y}^{k+1}=\left(\operatorname{fold}\left(\operatorname{unfold}\left(\mathcal{S}_{\frac{\mu}{\rho},w,p}(\mathcal{B}^{k}_\Phi)\right);\operatorname{unfold}\left(\mathcal{B}^{k}_{\Phi^c}\right)\right)\right)_{\overline{\Phi}^\top},
\end{equation}
where $\mathcal{B}^{k}=\mathcal{G}^{k+1}+\frac{\mathcal{C}^k}{\rho^k}$, $\mathcal{S}_{\frac{\mu}{\rho},w,p}(\mathcal{B}^{k}_\Phi)$ see the proof of the Theorem \ref{thm1} in the Appendix \ref{A1}.

Based on the description above, the pseudo code is summarized in Algorithm \ref{alg}.

\begin{algorithm}
	\caption{Solving problem (\ref{our})}
	\label{alg}
	\begin{algorithmic}
		\STATE{\textbf{Input:} $\left\{X_i\right\}_{i=1}^m$, $\left\{P_i\right\}_{i=1}^m$, $w$, $\mathcal{M}$, $\lambda$, $\mu$, $\gamma$.}
		\STATE{\textbf{Initialize:} $k=0$, $\eta=1.1$, $\rho^0=10^{-4}$, $\left\{\delta_i^0=\frac{1}{m}\right\}_{i=1}^m$, $\mathcal{C}^0=0$, $\mathcal{G}^0=\mathcal{Y}^0=\mathcal{M}$.}
		\WHILE{not converged}
		\STATE Update $A^{k+1}$ via Eq.(\ref{updateA}) 
		\STATE Update $W_i^{k+1}$ via $\left(X_iP_iX_i^\top\right)^{-1}X_iP_i{A^{k+1}}^\top$ for all $i$.
		\STATE  Update $\delta_i^{k+1}$ via definition for all $i$.
		\STATE Update $\mathcal{G}^{k+1}$ via Algorithm \ref{alg:g}.
		\STATE Update $\mathcal{Y}^{k+1}$ via Eq.(\ref{updateY2})
		\STATE Update $\mathcal{C}^{k+1}$ via $\mathcal{C}^{k+1}=\mathcal{C}^k+\rho^k\left(\mathcal{G}^{k+1}-\mathcal{Y}^{k+1}\right)$.
		\STATE Update $\rho^{k+1}$ via $\rho^{k+1}=\eta\rho^k$.
		\STATE $k\leftarrow k+1$.
		\ENDWHILE
		\STATE{\textbf{Output:} $A$, $\left\{W_i\right\}_{i=1}^m$}
	\end{algorithmic}
\end{algorithm}


\subsection{Convergence Analysis}
\label{sec:convergence}
We mainly analyze the convergence property of Algorithm \ref{alg} in this section. To prove the convergence of Algorithm \ref{alg}, we first prove the bounded of sequences $\left\{\mathcal{C}^k\right\}$, $\left\{\mathcal{Y}^k\right\}$ etc.. Lemma \ref{lemc} is given first and is proved in the Appendix \ref{A2}. Then convergence is shown as in Theorem \ref{thm2}, which is proved in the Appendix \ref{A3}.
\begin{lemma}
	\label{lemc}
	The sequence $\left\{\mathcal{C}_1^k\right\}$, $\left\{\mathcal{Y}^k\right\}$, $\left\{A^k\right\}$, $\left\{\left\{W_i^k\right\}_{i=1}^m\right\}$,  $\left\{\mathcal{G}^k\right\}$, and the augmented Lagrangian function Eq.(\ref{largfun}), which are generated by Algorithm \ref{alg}, are bounded.  
\end{lemma}

\begin{theorem}
	\label{thm2}
	The sequence $\left\{\mathcal{G}^k\right\}$ and $\left\{\mathcal{Y}^k\right\}$ generated by Algorithm satisfy
	\begin{enumerate}
		\item $\lim\limits_{k \to \infty}\left\|\mathcal{G}^k-\mathcal{Y}^k\right\|_F=0$,
		\item $\lim\limits_{k\to\infty}\left\|\mathcal{G}^{k+1}-\mathcal{G}^k\right\|_F=0$,
		\item $\lim\limits_{k\to\infty}\left\|\mathcal{Y}^{k+1}-\mathcal{Y}^k\right\|_F=0$.
	\end{enumerate}
\end{theorem}
\section{Experiments and Analyses}
\label{sec:exper}
\subsection{Experimental Settings}
\textbf{Datasets: }The four datasets used in the experiment are shown Table \ref{tab:datadis}, and their details are described below.
\begin{enumerate}
	\item \textit{Handwritten Digit Database}\footnote{http://archive.ics.uci.edu/ml/datasets/Multiple+Features}: This dataset contains ten categories, i.e., 1-10. Each category has 200 samples and a total of six views, i.e., pixel averages, Fourier coefficients, profile correlations, Zernike moment, Karhunen-love coefficient, and morphological. In this experiment, we only used the first five views. 
	\item \textit{3 Sources}\footnote{http://erdos.ucd.ie/datasets/3sources.html} This dataset contains 948 stories from three online resources, BBC, Reuters, and Guardian. In this experiment a subset containing 169 stories were reported in all resources was used for comparison with other methods. These 169 stories were classified into six classes, which are business, entertainment, health, politics, sport, and technology.
	\item \textit{BBC Sport}\cite{PSPDKDC} This dataset contains 737 documents about sport news articles. These documents are described by 2-4 views and divided into 5 categories. In our experiments, we selected a subset containing 116 samples described by four views.
	\item \textit{Orl}\footnote{http://cam-orl.co.uk/facedatabase.html}: This is a very popular face dataset with 400 images of faces provided by 40 persons, each with 10 photos. Since Orl is a single-view dataset, we extracted LBP, GIST, and pyramid of the histogram of oriented gradients and combined them with pixel features to form a 4-views dataset.
\end{enumerate}

\begin{table}[htbp]
	\small
	\centering
	\caption{Description of the multi-view datasets}
	\begin{tabular}{ccccc}
		\toprule
		Dataset     & Class & View & Samples & Features         \\ \midrule
		Handwritten& 10    & 5    & 2000    & 240/76/216/47/64 \\ 
		3 Sources  &  6  &  3  &  169  &  3560/3631/3036 \\
		BBC Sport &  5  &  4  &  116  &  1991/2063/2113/2158\\
		Orl& 40      & 4     & 400        & 41/399/399/41  \\ 
		\bottomrule
	\end{tabular}
	\label{tab:datadis}
\end{table}

\textbf{Incomplete multi-view data construction: }For each of the above datasets, we randomly selected $p\%$ ($p \in \left\{90, 70, 50, 30\right\}$) samples to make them into incomplete samples, where the deletion of randomly selected instances generates the incomplete samples.

\textbf{Compared methods: }The following methods can deal with incomplete multi-view data to be used as baselines in this experiment.
\begin{enumerate}
	\item BSV \cite{IMMOVDG}: BSV performs $k$-means clustering and nearest neighbor classification on all views and reports the optimal clustering and classification results, where the missing instances are completed in the average instance of the corresponding view. 
	\item Concat \cite{IMMOVDG}: Concatenating all views into one single view and exploiting $k$-means and nearest neighbor to obtain the final clustering and classification results, where the missing instances are also filled	in the average instance like BSV.
	\item DAIMC \cite{DAIMC}: This method used two techniques for consistent representation of all views, namely matrix decomposition based on alignment of instance information and sparse regression based on basis matrices.
	\item IMSC\_AGL \cite{wenjieIMSCAG}: The method acquires consensus low-dimensional representation based on spectral clustering and constructed graphs from low-rank representation.
	\item TCIMC \cite{TCBIMC}: This method obtains the complete graph by tensor recovery for each view, and differ from our method in the tenor norm. 
\end{enumerate}

Our method firstly obtains the low-dimensional representation $A$ of all samples and then uses $k$-means on $A$. We preform $k$-means several times to get the average result since it is sensitive to the initial value. We use $k$-nearest-neighbor to construct a graph for each view, i.e.,
\begin{equation}
	\label{graph}
	M_{i(k,l)}=\left\{\begin{split}
		&e^{-\frac{\left\|X_{i\left(k,:\right)}-X_{i\left(l,:\right)}\right\|_2^2}{2\sigma^2}}, \quad X_{i\left(k,:\right)}\in k\text{-NN}\left(X_{i\left(l,:\right)}\right)\\
		&0, \qquad\qquad\qquad\qquad\text{otherwise} 
	\end{split} \right.,
\end{equation}
where $k=10$, $\sigma=1$. For the missing positions in the $M_i$, we fill them with $0$. The $p$, $w$, and $\Phi$ in the tensor norm $\left\|\cdot\right\|_{\Phi,w,S_p}$ are set $0.6$, $\textbf{1}$, and discrete cosine transform, respectively.

\textbf{Evaluation metrics: }The clustering of the experiment in this paper uses three common metrics, i.e., clustering accuracy (ACC), normalized mutual information (NMI), and Purity.

\subsection{Experimental Result}
	Table \ref{tab:clusterresult} shows the clustering results of the six methods on four datasets. We can draw the following conclusions from it.
\begin{enumerate}
	\item Our method has the best results in most cases compared to the other five methods. In the clustering task, the proposed method is approximately $5\%$ better than the second-best method on the Orl dataset. Inevitable, some data for which some measure is not optimal, e.g., the clustering task of Handwritten, 3Sources, and BBC Sport datasets.
	\item IMSC\_AGL, TCIMC, and our method are graph-based methods. Table \ref{tab:clusterresult} shows that IMSC\_AGL, TCIMC, and our method have better clustering performance than DAIMC on all datasets. This indicates that graph tools can improve the clustering performance. Both TCIMC and our method are based on graph complementation and both are based on tensor recovery, the difference is that different tensor norm are used. The experiments show that our method has better performance in clustering than TCIMC in most datasets.
	\item From the experimental result in Table \ref{tab:clusterresult}, it can be seen that the clustering performance have a downward trend in most datasets overall with the increase of view missing rate. This phenomenon shows that it is difficult to learn a reasonable and effective consensus representation on a dataset with a high miss rate. Too many missing instances can result in insufficient consensus and complementarity among views. The experiments results on all datasets show that our proposed method has a greater advantage in the case of higher missing rate.
\end{enumerate}

\begin{sidewaystable}[htbp]
	\tiny
	\centering
	\caption{Average and standard values of ACC (\%), NMI(\%), AND Purity(\%) of seven methods over four datasets.}
	\begin{tabular}{cccccccccccccc}
		\toprule    \multirow{2}[4]{*}{Datasets} & \multirow{2}[4]{*}{Method\textbackslash{}Rate} & \multicolumn{4}{c}{ACC}       & \multicolumn{4}{c}{NMI}       & \multicolumn{4}{c}{Purity} \\
		\cmidrule{3-14}          &       & 0.9   & 0.7   & 0.5   & 0.3   & 0.9   & 0.7   & 0.5   & 0.3   & 0.9   & 0.7   & 0.5   & 0.3 \\
		\midrule
		\multirow{6}[2]{*}{Handewritten} & BSV   & 47.15±2.86 & 52.04±24.95 & 62.51±19.99 & 67.47±35.51 & 38.79±1.09 & 45.34±9.94 & 54.09±8.35 & 60.78±11.11 & 47.15±2.83 & 52.77±17.79 & 62.51±17.69 & 68.80±23.08 \\
		& Concat & 39.96±3.37 & 44.37±10.75 & 48.95±2.88 & 53.12±5.01 & 32.91±1.51 & 37.91±4.83 & 43.27±1.73 & 49.02±1.12 & 40.65±1.71 & 46.22±4.08 & 51.27±1.21 & 57.06±2.66 \\
		& DAIMC & 84.51±0.00 & 85.21±0.06 & 87.13±0.04 & 86.61±0.05 & 73.23±0.01 & 76.64±0.01 & 77.17±0.03 & 77.24±0.02 & 84.51±0.01 & 86.75±0.01 & 87.18±0.02 & 86.73±0.23 \\
		& IMSC\_AGL & 94.14±0.00 & 94.32±0.00 & \textbf{95.40±0.00} & \textbf{96.04±0.00} & 88.07±0.00 & 88.12±0.01 & 90.01±0.00 & 91.34±0.00 & 94.14±0.00 & 94.34±0.00 & \textbf{95.40±0.00} & \textbf{96.04±0.00} \\
		& TCIMV & 86.79±0.98 & 86.35±1.26 & 85.90±0.36 & 85.15±1.01 & 88.95±1.34 & 89.77±0.54 & 89.8±2.36 & 89.58±0.95 & 88.01±1.32 & 88.40±2.01 & 88.40±0.66 & 88.10±1.32 \\
		& Our   & \textbf{99.60±0.00} & \textbf{99.95±0.00} & 86.08±0.63 & 86.95±0.00 & \textbf{98.94±0.00} & \textbf{99.86±0.00} & \textbf{90.10±2.13} & \textbf{91.37±0.00} & \textbf{99.60±0.00} & \textbf{99.95±0.00} & 88.58±0.62 & 89.00±0.00 \\
		\midrule
		\multirow{6}[2]{*}{3Sources} & BSV   & 35.50±3.11 & 35.74±12.46 & 36.09±8.01 & 35.62±15.77 & 8.53±11.81 & 9.29±12.87 & 8.56±11.76 & 7.92±6.29 & 38.99±5.24 & 39.88±12.93 & 39.64±16.10 & 39.17±12.2 \\
		& Concat & 37.33±6.96 & 38.04±30.50 & 40.35±47.45 & 46.39±35.41 & 7.96±14.37 & 11.36±69.01 & 13.87±75.81 & 22.97±98.79 & 39.40±18.61 & 42.48±61.37 & 45.32±74.47 & 52.60±61.97 \\
		& DAIMC & 46.86±0.00 & 53.37±1.93 & 50.29±0.00 & 49.76±0.04 & 44.29±0.00 & 46.03±3.02 & 45.38±0.16 & 47.53±0.35 & 62.13±0.00 & 68.52±0.68 & 68.28±0.10 & 67.57±0.06 \\
		& IMSC\_AGL & 60.95±1.55 & 57.45±1.20 & \textbf{65.62±5.94} & \textbf{68.11±2.83} & 59.17±0.19 & 60.55±0.10 & \textbf{57.89±6.65} & \textbf{61.18±0.86} & 77.75±0.09 & 77.28±0.09 & \textbf{76.56±6.08} & 76.69±1.26 \\
		& TICMV & 68.64±0.00 & 69.23±0.00 & 56.80±0.00 & 69.82±0.00 & 63.85±0.00 & 54.34±0.02 & 52.50±0.00 & 57.98±0.00 & 79.23±0.00 & 75.14±0.00 & 72.19±0.00 & \textbf{78.10±0.00} \\
		& Our   & \textbf{74.41±0.16} & \textbf{78.70±0.14} & 59.76±0.06 & 59.16±0.08 & \textbf{63.97±0.22} & \textbf{66.29±0.11} & 52.47±1.14 & 60.20±0.20 & \textbf{79.28±0.09} & \textbf{82.25±0.14} & 70.41±0.06 & 74.56±0.00 \\
		\midrule
		\multirow{6}[2]{*}{BBC Sport} & BSV   & 33.88±4.46 & 34.13±7.63 & 35.26±4.86 & 34.655±14.00 & 5.82±3.72 & 6.23±5.19 & 6.11±10.06 & 6.54±19.41 & 35.00±3.34 & 35.17±6.73 & 36.21±5.11 & 35.60±13.71 \\
		& Concat & 33.36±4.79 & 34.91±21.35 & 33.62±3.30 & 37.58±61.13 & 5.81±2.24 & 6.27±7.76 & 5.37±5.26 & 11.01±63.76 & 34.65±2.61 & 36.55±19.19 & 35.29±4.03 & 39.14±55.19 \\
		& DAIMC & 60.69±0.19 & 66.38±1.98 & 61.81±1.33 & 62.16±5.85 & 46.45±0.95 & 55.52±0.55 & 45.28±2.29 & 37.05±19.07 & 68.10±0.83 & 75.26±1.49 & 68.62±1.51 & 65.00±7.46 \\
		& IMSC\_AGL & 66.21±0.13 & 78.45±0.00 & 69.66±0.79 & 75.00±0.00 & 51.17±0.83 & 69.55±0.00 & 57.68±0.31 & 68.60±0.40 & 76.55±0.13 & 87.93±0.00 & 82.58±0.79 & 87.84±0.07 \\
		& TICMV & 81.03±0.00 & 81.62±0.00 & 79.31±0.00 & \textbf{81.89±0.00} & 78.20±0.00 & \textbf{83.58±0.00} & 71.07±0.00 & \textbf{76.19±0.00} & 91.03±0.00 & 93.82±0.00 & 88.79±0.00 & 90.21±0.00 \\
		& Our   & \textbf{83.62±0.00} & \textbf{81.90±0.00} & \textbf{82.76±0.00} & 80.17±0.00 & \textbf{83.03±0.00} & 79.91±0.00 & \textbf{77.31±0.00} & 74.50±0.00 & \textbf{93.11±0.00} & \textbf{93.96±0.00} & \textbf{92.24±0.00} & \textbf{90.52±0.00} \\
		\midrule
		\multirow{6}[2]{*}{Orl} & BSV   & 43.10±2.98 & 40.70±1.22 & 42.27±1.92 & 42.4±1.6 & 47.81±0.20 & 47.93±0.80 & 48.58±0.79 & 48.62±0.13 & 46.18±1.20 & 44.47±0.42 & 45.92±1.04 & 45.72±0.17 \\
		& Concat & 16.08±0.88 & 15.22±2.23 & 16.87±2.88 & 21.50±135.6 & 24.28±2.91 & 24.72±5.56 & 29.04±9.52 & 35.14±178.1 & 19.10±0.57 & 18.25±2.09 & 19.82±2.97 & 24.90±146.8 \\
		& DAIMC & 56.18±6.47 & 58.7±4.36 & 61.08±1.89 & 65.05±5.19 & 70.98±1.91 & 73.70±2.07 & 76.18±1.89 & 79.90±2.60 & 58.82±4.76 & 61.05±4.71 & 63.70±3.15 & 68.15±4.96 \\
		& IMSC\_AGL & 71.20±4.95 & 75.58±4.16 & 71.45±2.02 & 73.85±4.44 & 83.22±1.39 & 86.79±0.36 & 84.57±0.57 & 86.10±0.57 & 74.13±3.80 & 77.90±1.86 & 74.12±2.18 & 76.75±2.19 \\
		& TCIMV & 86.50±1.06 & 84.95±0.85 & 85.45±0.54 & 84.00±2.04 & 91.06±0.09 & 92.94±0.07 & 92.27±0.17 & 91.24±0.41 & 87.40±0.09 & 88.35±0.03 & 87.35±0.17 & 86.60±0.41 \\
		& Our   & \textbf{92.10±2.04} & \textbf{88.72±4.83} & \textbf{87.87±2.37} & \textbf{89.45±1.76} & \textbf{95.02±0.24} & \textbf{95.25±0.36} & \textbf{94.98±0.16} & \textbf{96.23±0.28} & \textbf{92.65±1.07} & \textbf{90.63±1.88} & \textbf{90.07±0.75} & \textbf{92.30±1.27} \\
		\bottomrule
	\end{tabular}%
	\label{tab:clusterresult}%
\end{sidewaystable}%

\subsection{Analysis of the Parameter Sensitivity}
\begin{figure}[htb]
	\centering
	\subfigure[]{
		\label{figpsla} 
		\includegraphics[width=0.23\textwidth]{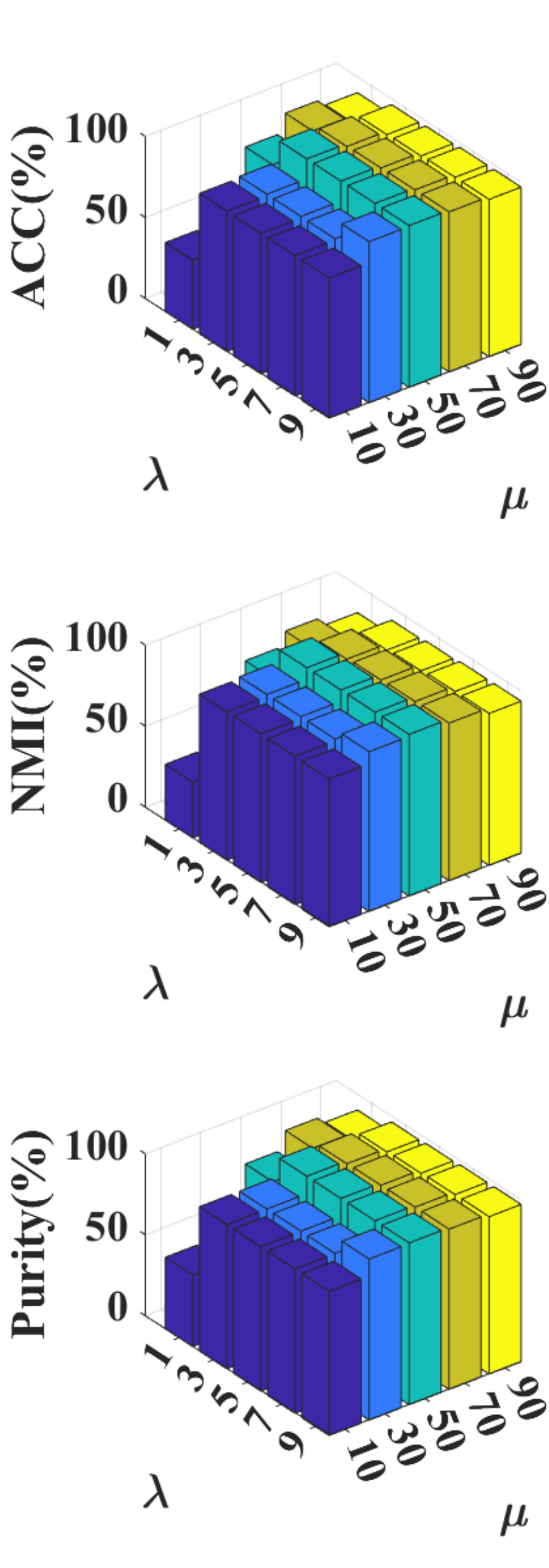}}
	\subfigure[]{
		\label{figpslb} 
		\includegraphics[width=0.23\textwidth]{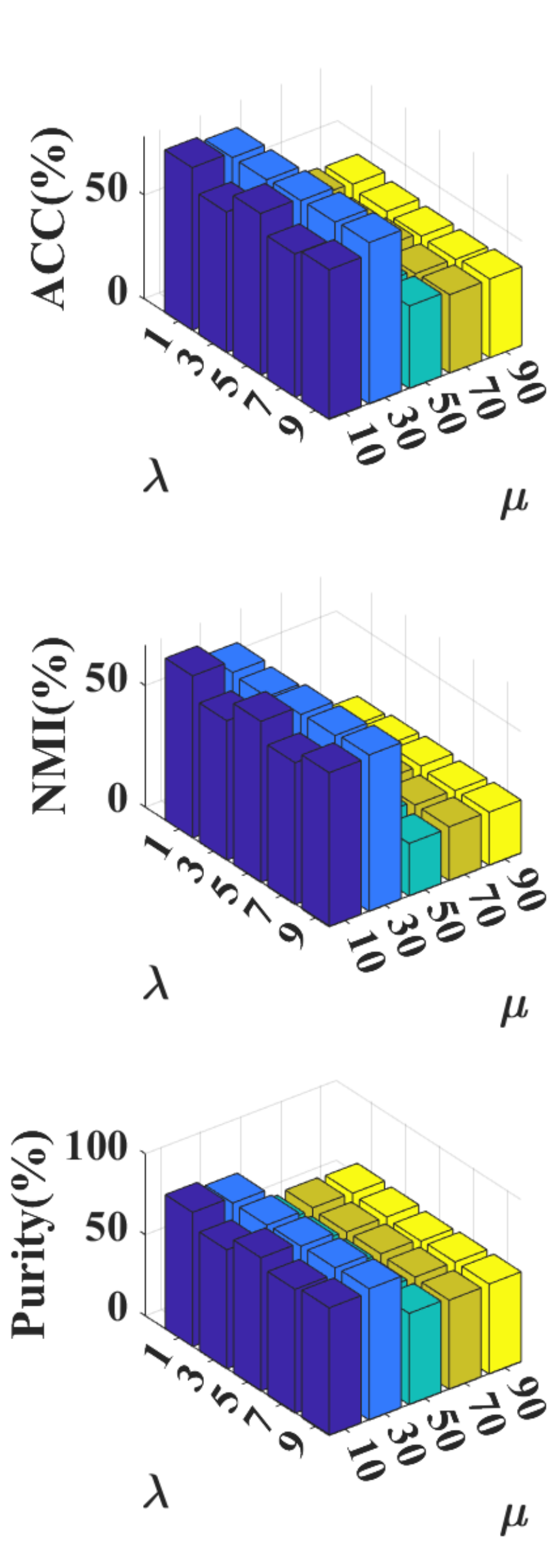}}
	\subfigure[]{
		\label{figpslc} 
		\includegraphics[width=0.23\textwidth]{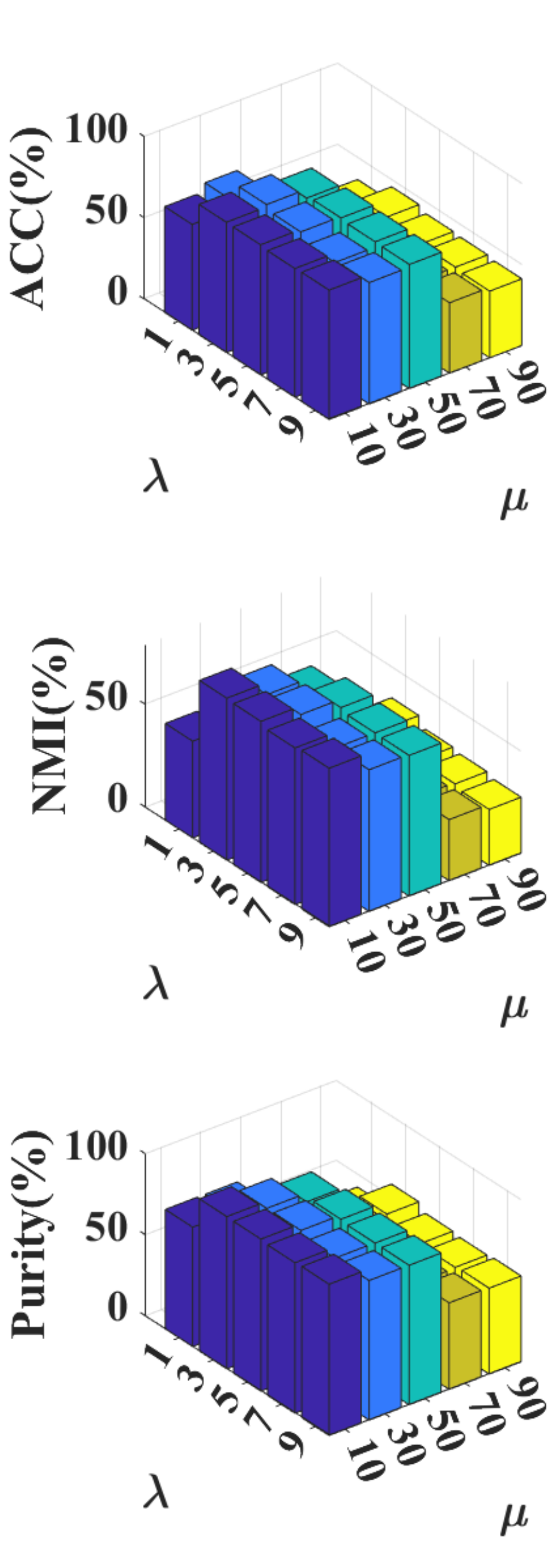}}
	\subfigure[]{
		\label{figpsld} 
		\includegraphics[width=0.23\textwidth]{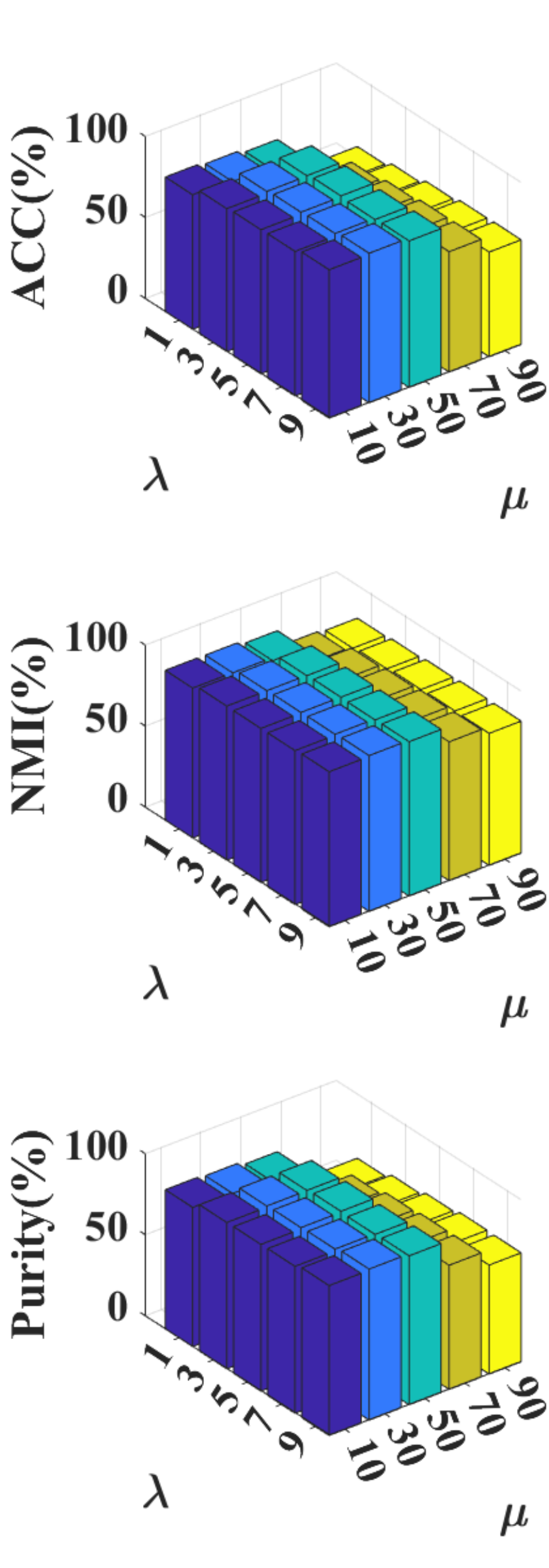}}
	\caption{The clustering performance of parameters $\lambda$ and $\mu$ in different combination on (a)	Handwritten, (b) 3Sources, (c) BBC Sport, and (d) Orl datasets with a missing rate of $70\%$.}
	\label{figparsenl} 
\end{figure}

\begin{figure}[htb]
	\centering
	\subfigure[]{
		\label{figpsa} 
		\includegraphics[width=0.48\textwidth]{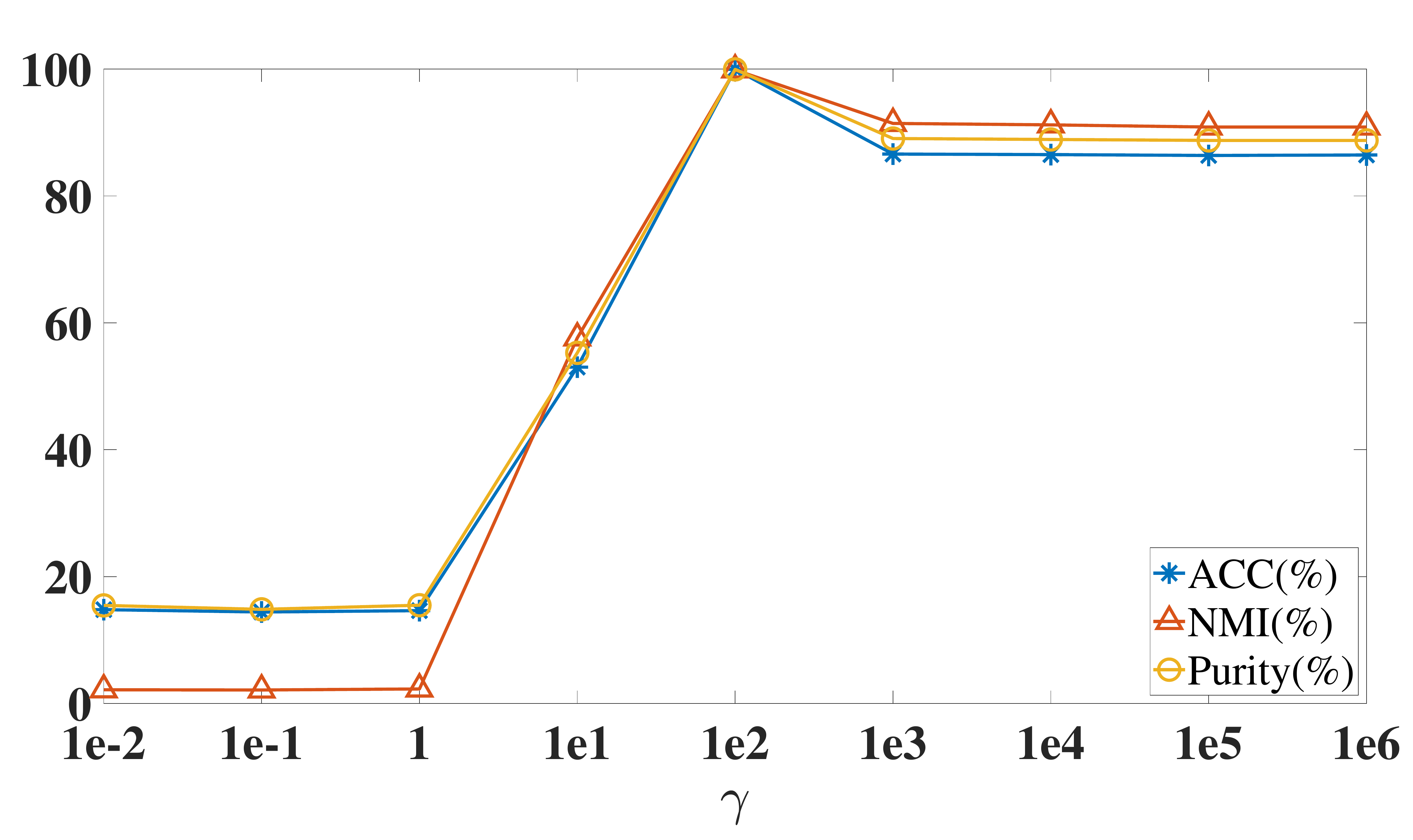}}
	\subfigure[]{
		\label{figpsb} 
		\includegraphics[width=0.48\textwidth]{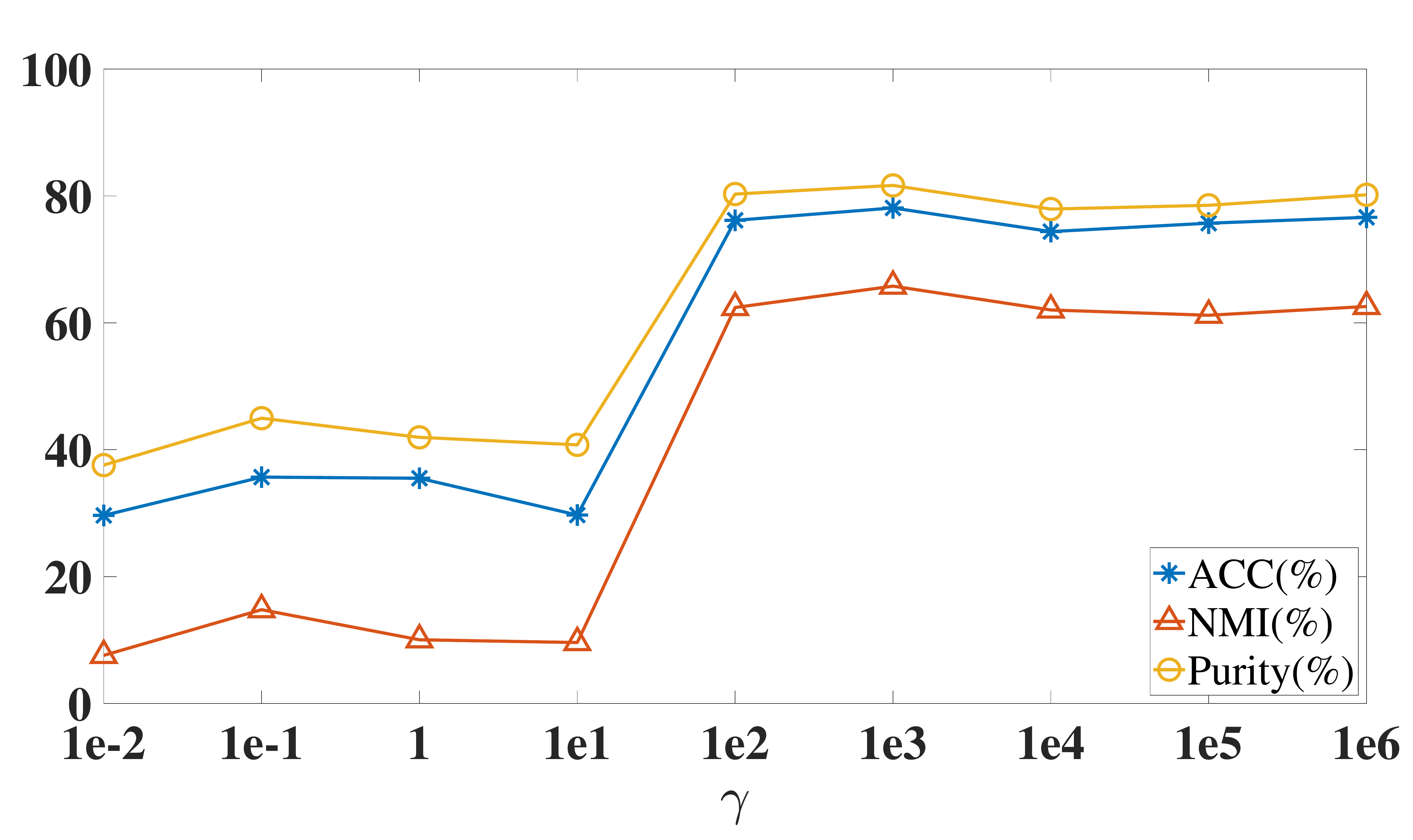}}\\
	\subfigure[]{
		\label{figpsc} 
		\includegraphics[width=0.48\textwidth]{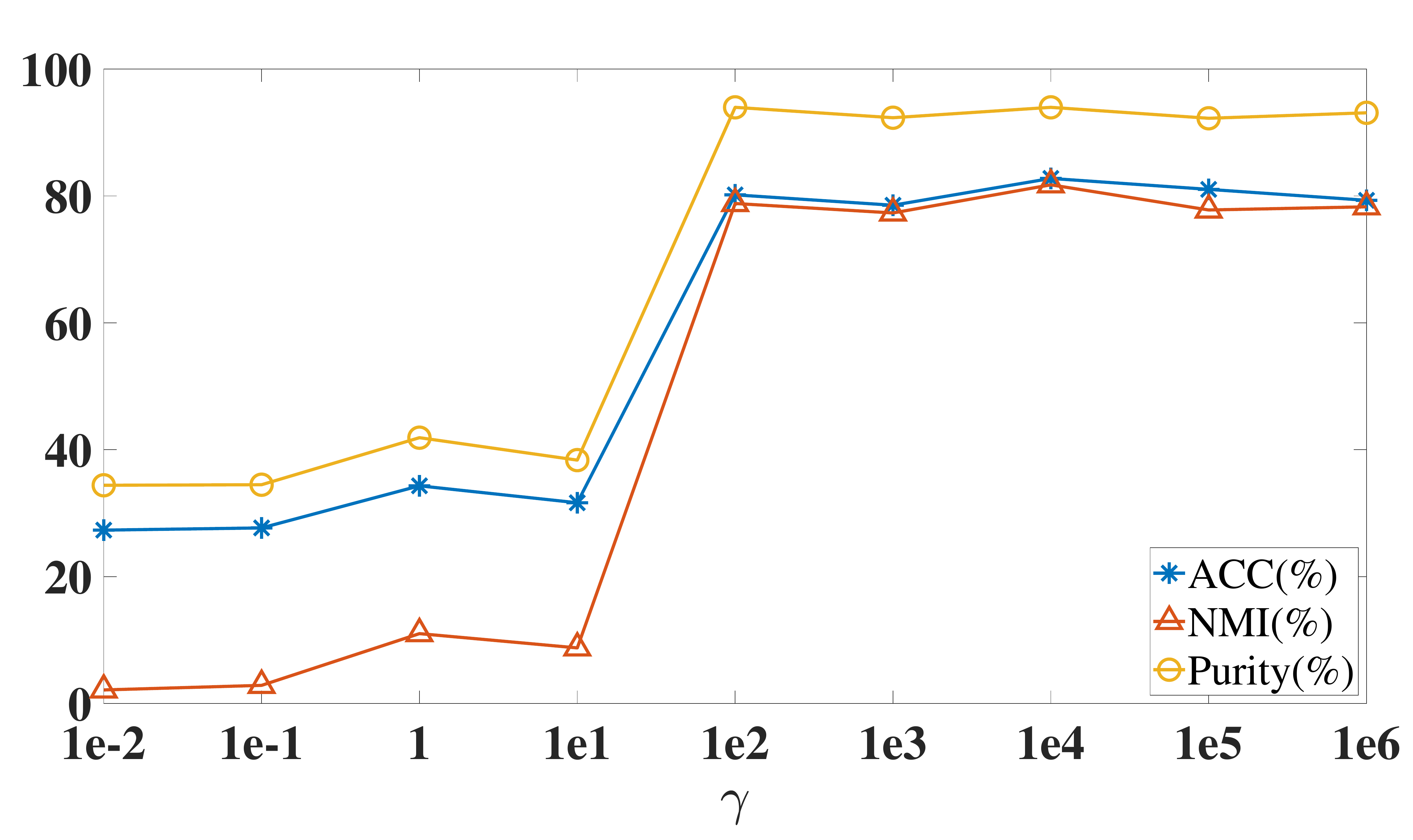}}
	\subfigure[]{
		\label{figpsd} 
		\includegraphics[width=0.48\textwidth]{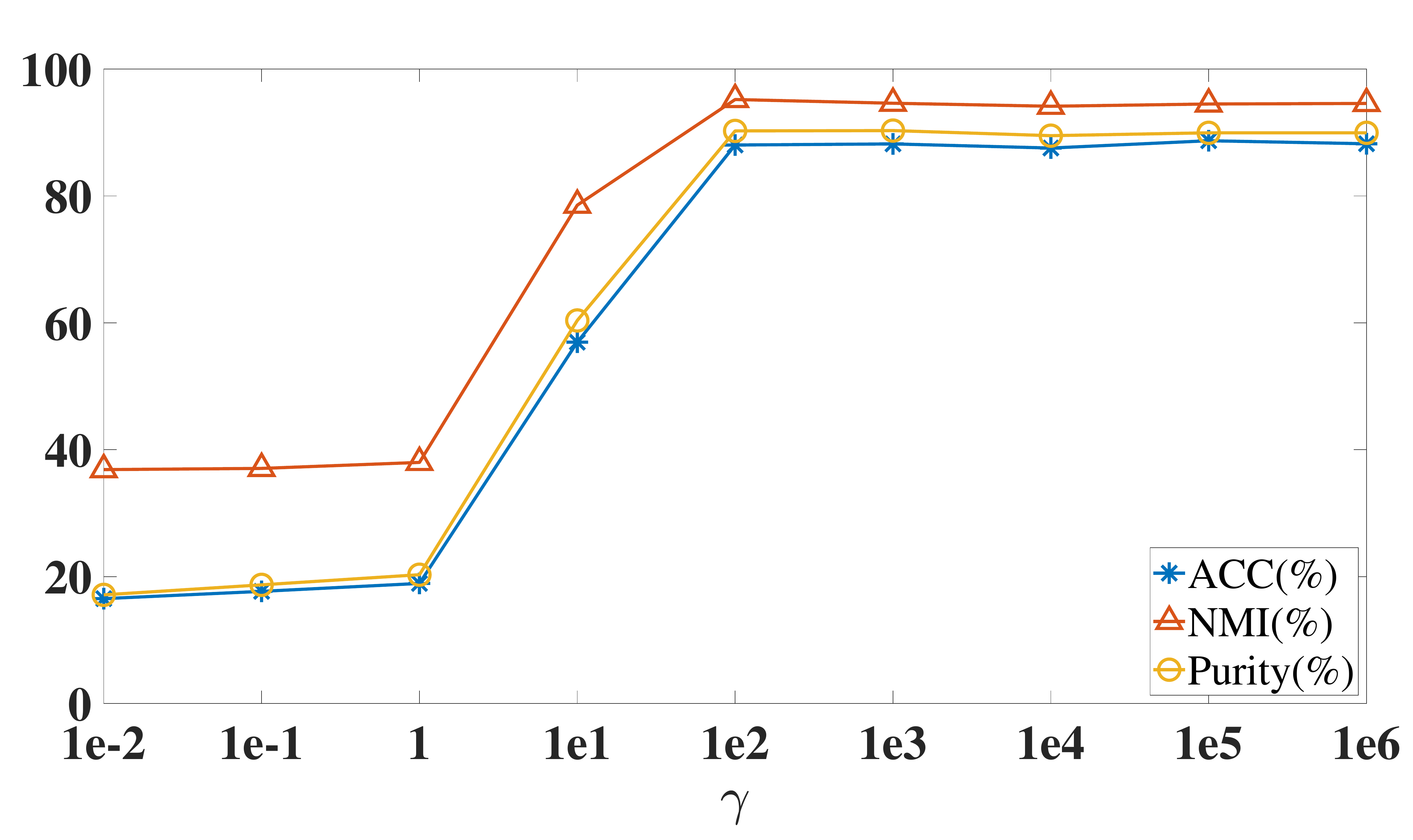}}
	\caption{The clustering performance of parameter $\gamma$ in different value on (a) Handwritten,
		(b) 3Sources, (c) BBC Sport, and (d) Orl datasets with a missing rate of $70\%$.}
	\label{figparseng} 
\end{figure}

This section analyzes the sensitive of three parameters $\lambda$, $\mu$, and $\gamma$ on Handwritten, 3Sources, BBC Sport, and Orl datasets. We first set up a candidate set for $\lambda$ and $\mu$, i.e. $\left\{1,3,5,7,9\right\}$ and $\left\{10,30,50,70,90\right\}$, then execute our proposed method for clustering with different combinations of two parameters ($\lambda\&\mu$). Then we also execute proposed method for clustering with different $\gamma$ on candidate set $\{1e-2,1e-1,1,1e1,1e2,1e3,1e4,1e5,1e6\}$. Fig.\ref{figparsenl} and Fig.\ref{figparseng} show the clustering performance versus the scale parameter $\lambda\&\mu$ and $\gamma$ on the four datasets with a missing rate of $70\%$ respectively. The experimental results show that our method is relatively stable for hyperparameters $\lambda$, $\mu$, and $\gamma$, and the clustering performance is usually optimal for $\lambda = 3$, $\mu=10$, and $\gamma =100$.

\subsection{Convergence Analysis}
\begin{figure}[htb]
	\centering
	\subfigure[]{
		\label{fig3a} 
		\includegraphics[width=0.48\textwidth]{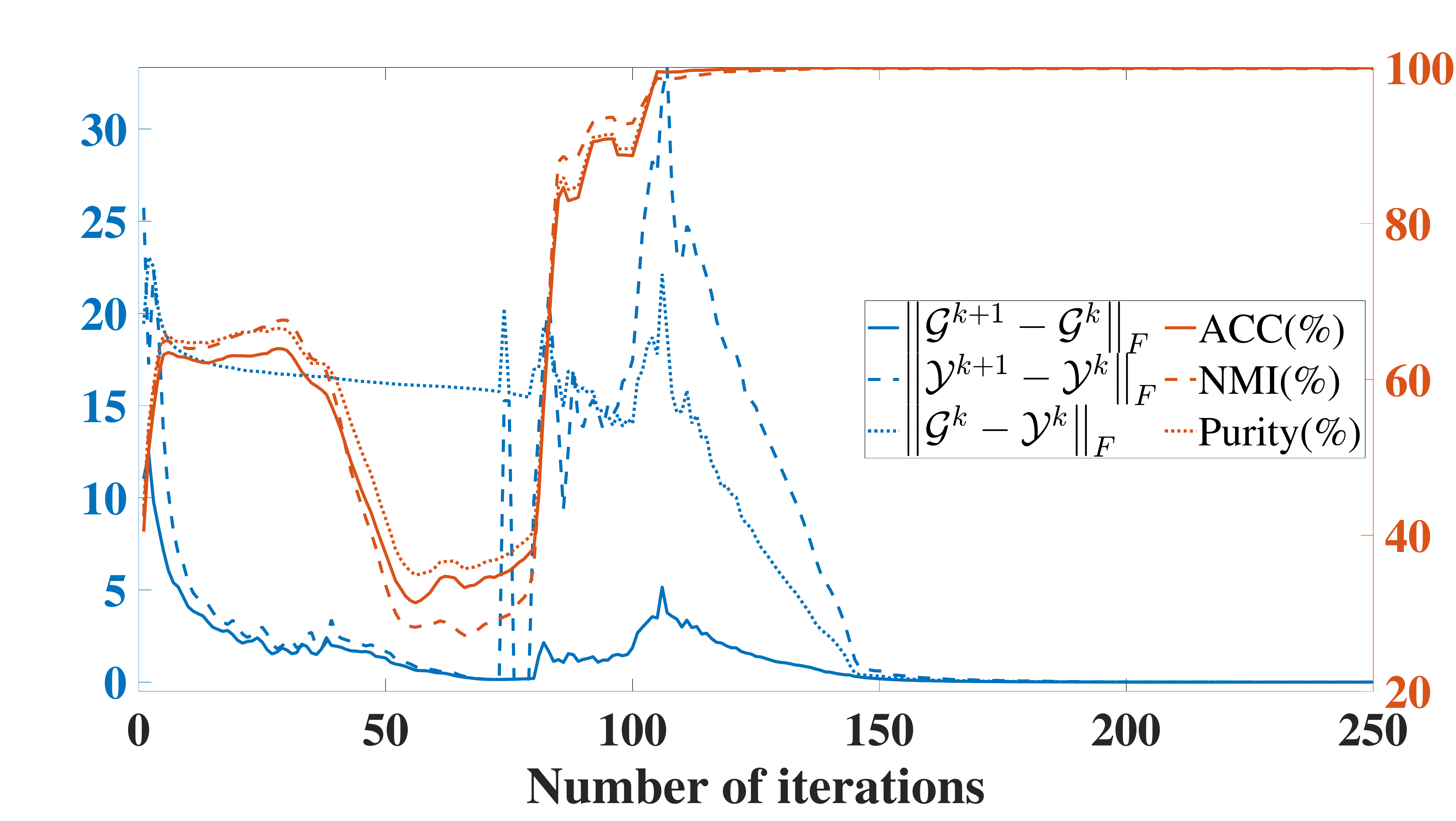}}
	\subfigure[]{
		\label{fig3b} 
		\includegraphics[width=0.48\textwidth]{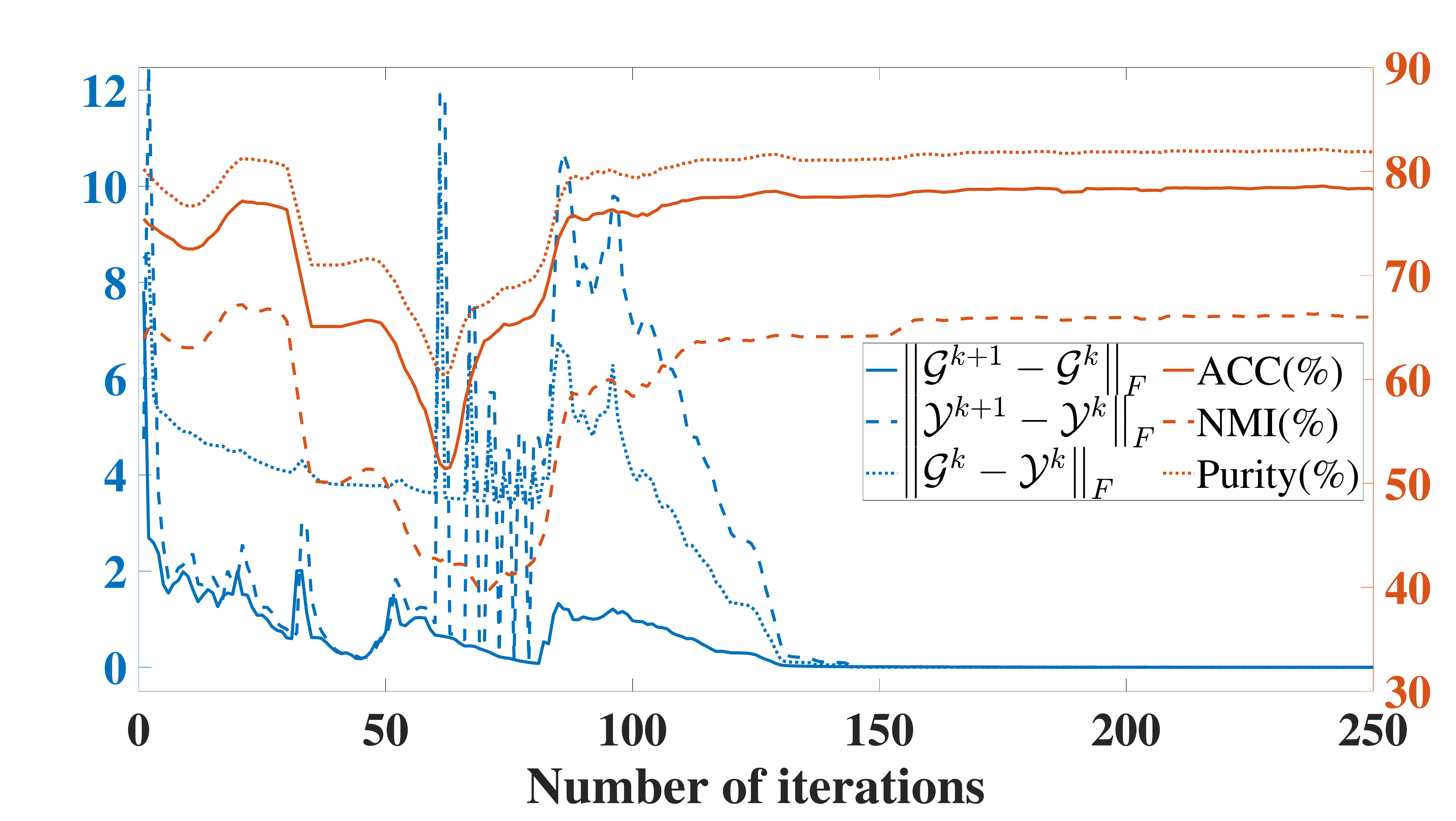}}\\
	\subfigure[]{
		\label{fig3c} 
		\includegraphics[width=0.48\textwidth]{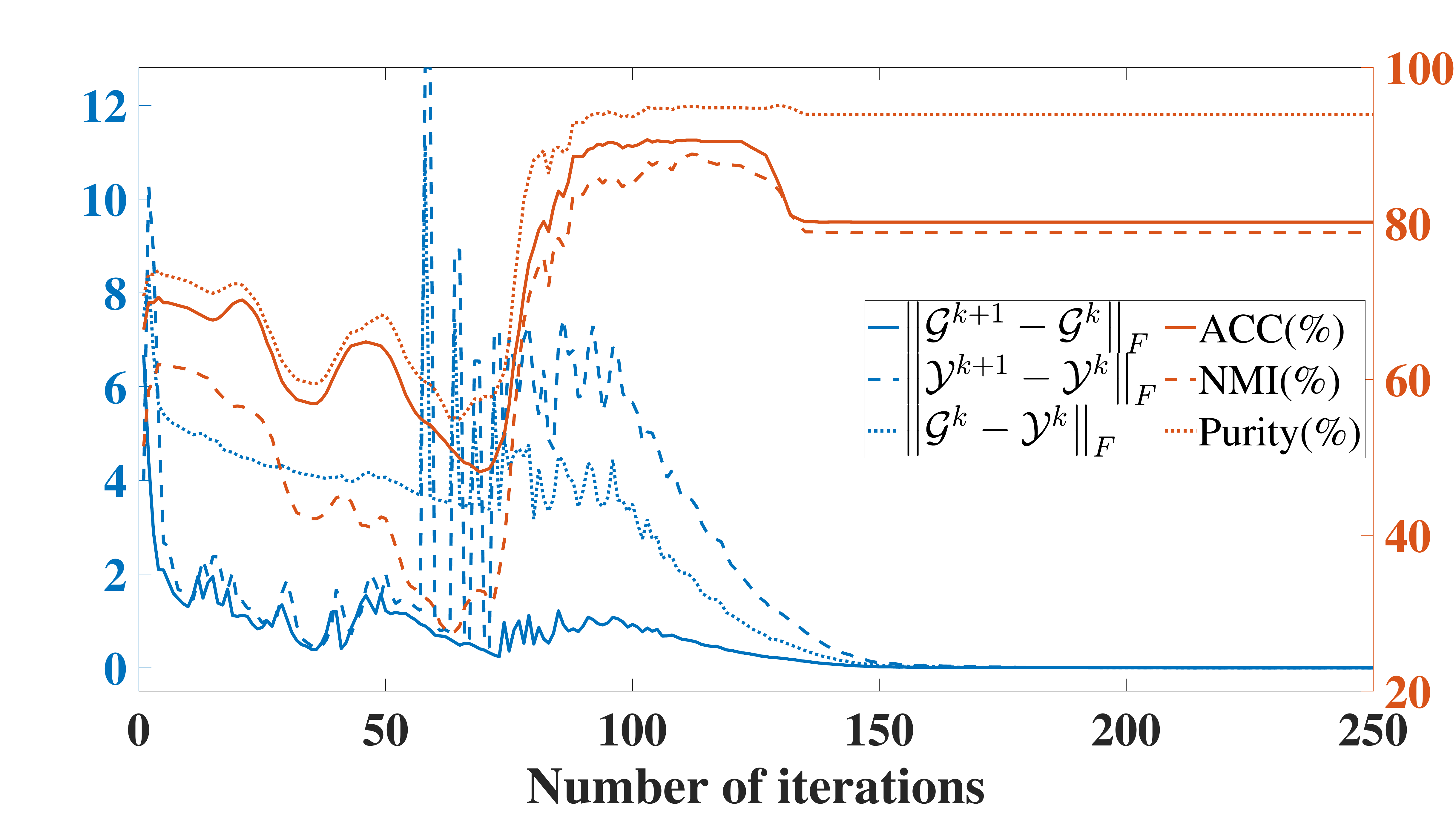}}
	\subfigure[]{
		\label{fig3d} 
		\includegraphics[width=0.48\textwidth]{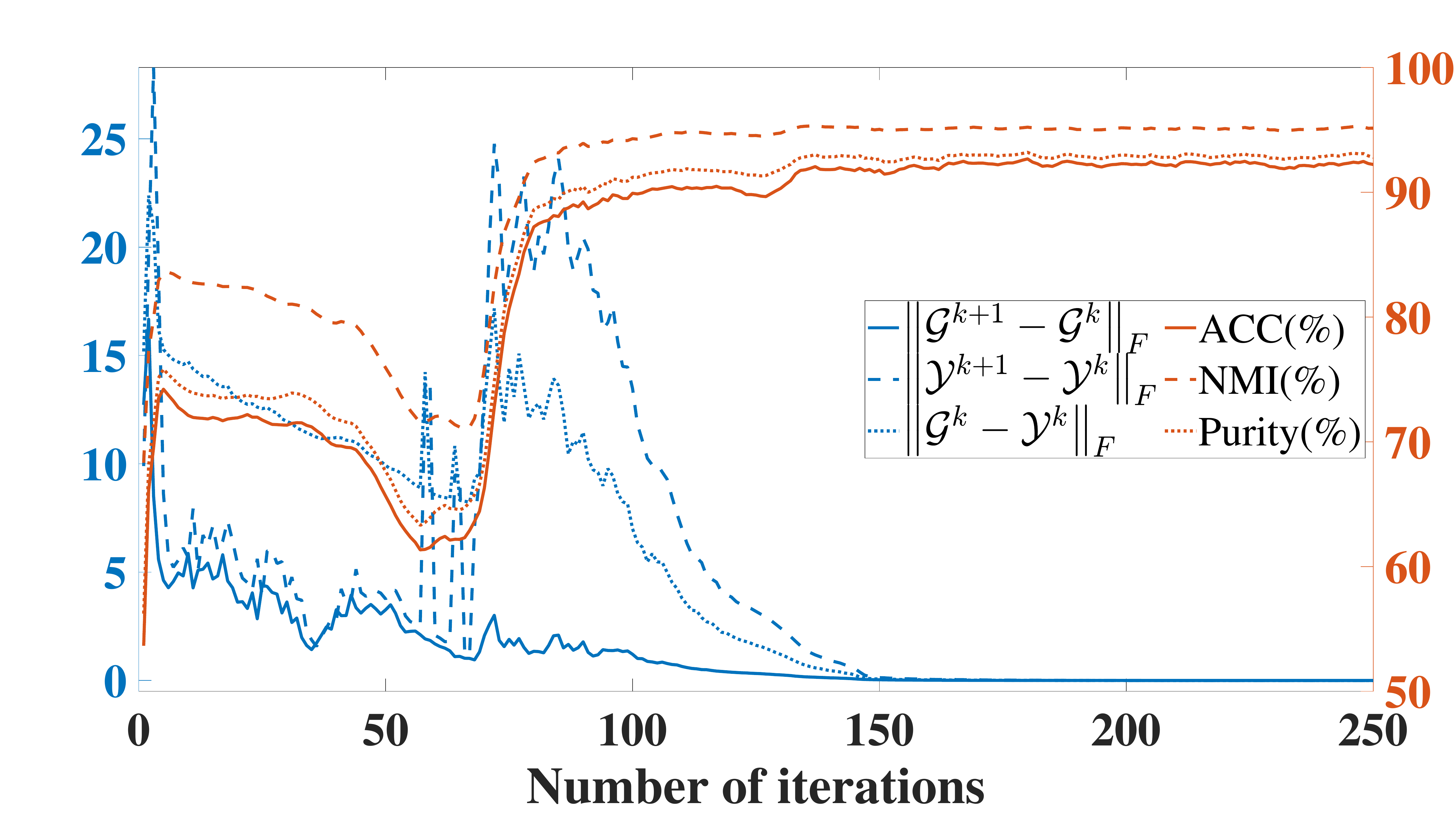}}
	\caption{The convergence of the algorithm and the clustering performance of the model with the number of iterations on (a) Handwritten, (b) 3 Sources, (c) BBC Sport, (d) Orl datasets with a missing rate of $70\%$.}
	\label{figconvergence} 
\end{figure}

In section \ref{sec:convergence}, we show that Algorithm \ref{alg} is convergent. Then we use numerical experiments to verify the convergence of the proposed algorithm. We conducted experiments on Handwritten, 3 Sources, BBC Sport, and Orl datasets with a missing rate of $70\%$. Figure \ref{figconvergence} shows the convergence of the algorithm and the clustering performance of the model as the number of iteration increases. It can be seen from Fig.\ref{figconvergence} that the optimization method provided by us has good convergence, in which $\left\|\mathcal{G}^{k+1}-\mathcal{G}^k\right\|_F$, $\left\|\mathcal{Y}^{k+1}-\mathcal{Y}^k\right\|_F$, and $\left\|\mathcal{G}^{k}-\mathcal{Y}^k\right\|_F$ can rapidly decline and tend to be $0$ in all four datasets. The clustering results also stabilize with increasing number of iterations.

\subsection{The Effect of the Parameter $r$}
\label{norm_nn}
\begin{figure}[h]
	\centering
	\subfigure[]{
		\label{figna} 
		\includegraphics[width=0.48\textwidth]{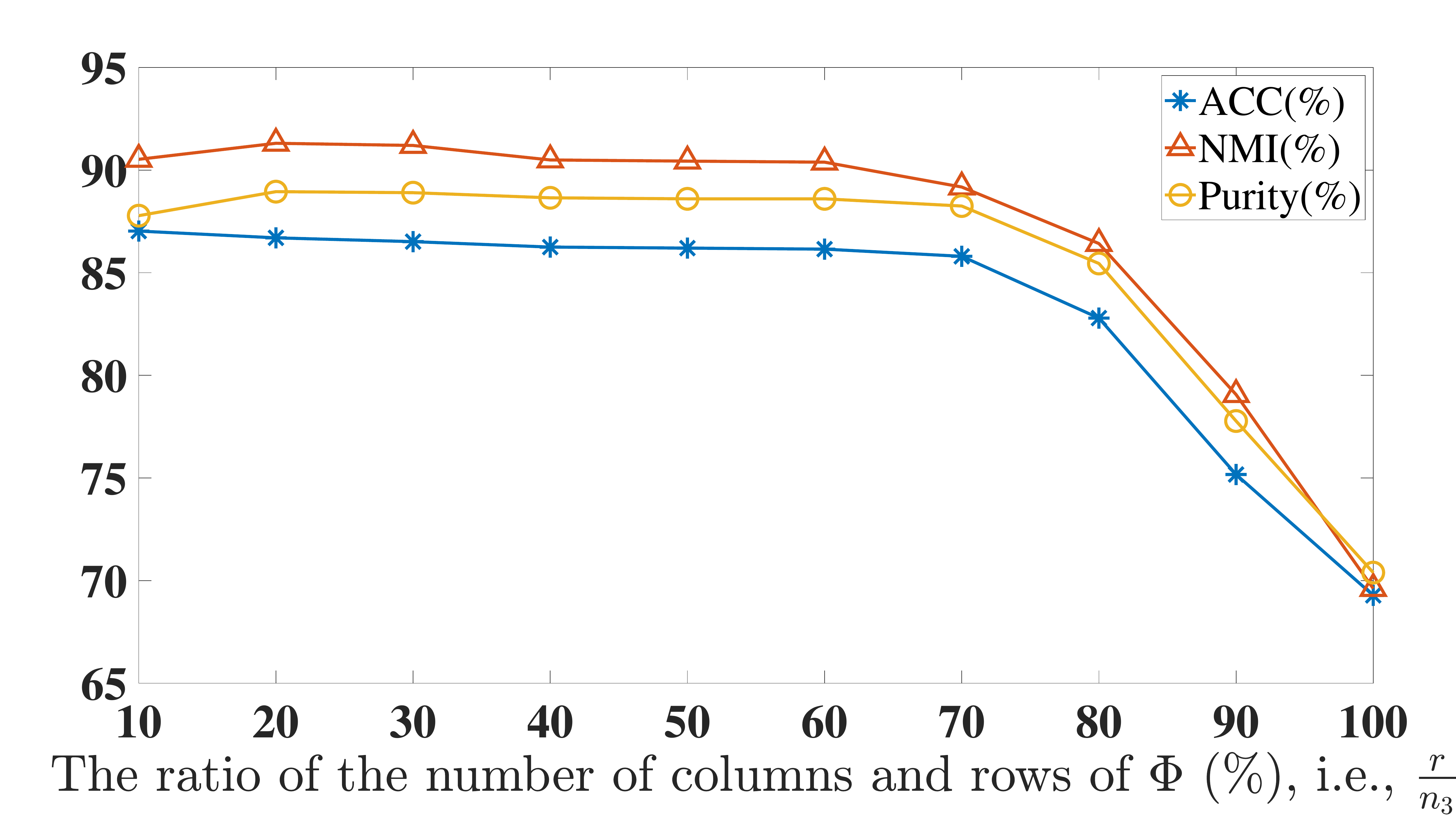}}
	\subfigure[]{
		\label{fignb} 
		\includegraphics[width=0.48\textwidth]{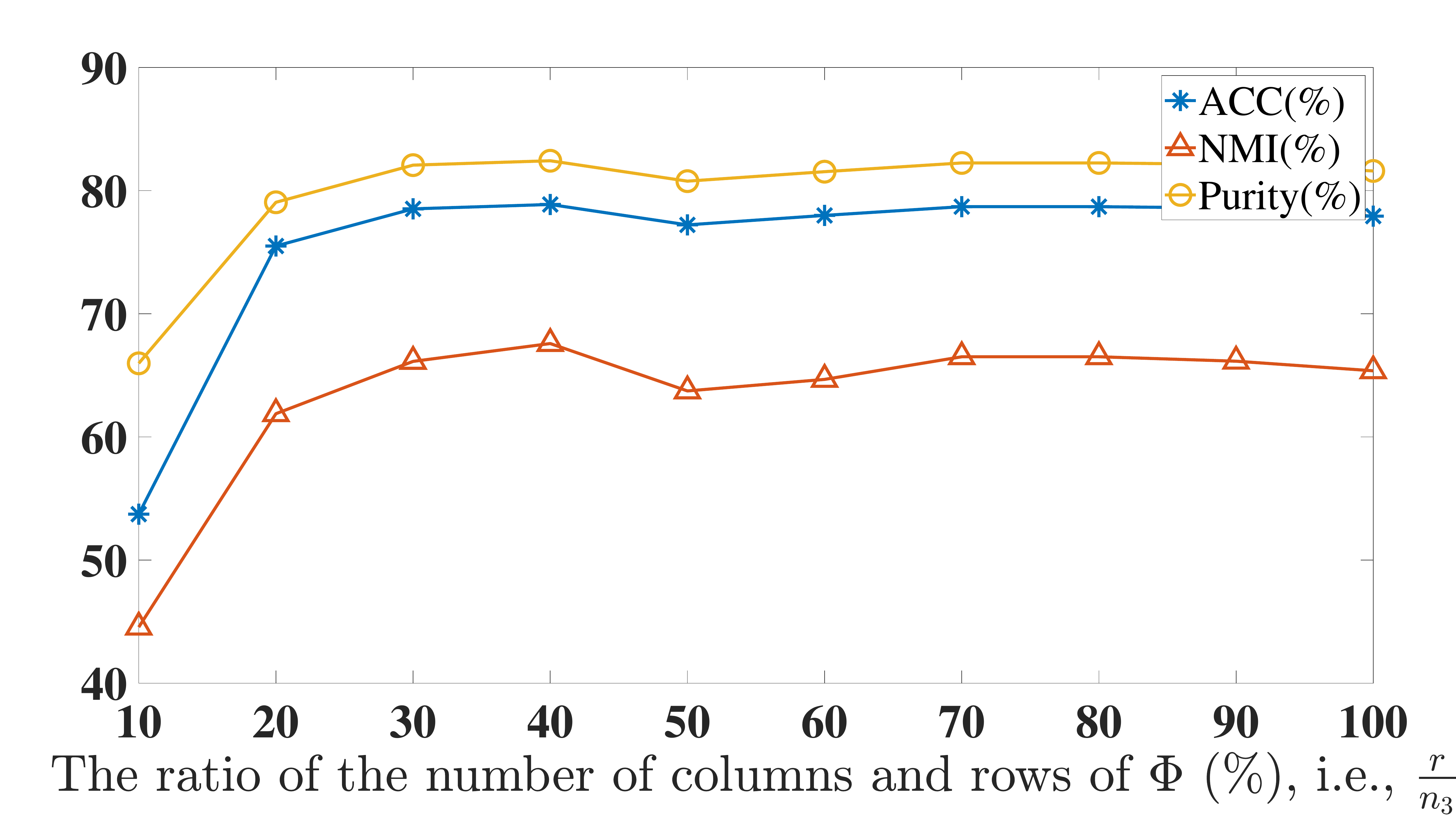}}\\
	\subfigure[]{
		\label{fignc} 
		\includegraphics[width=0.48\textwidth]{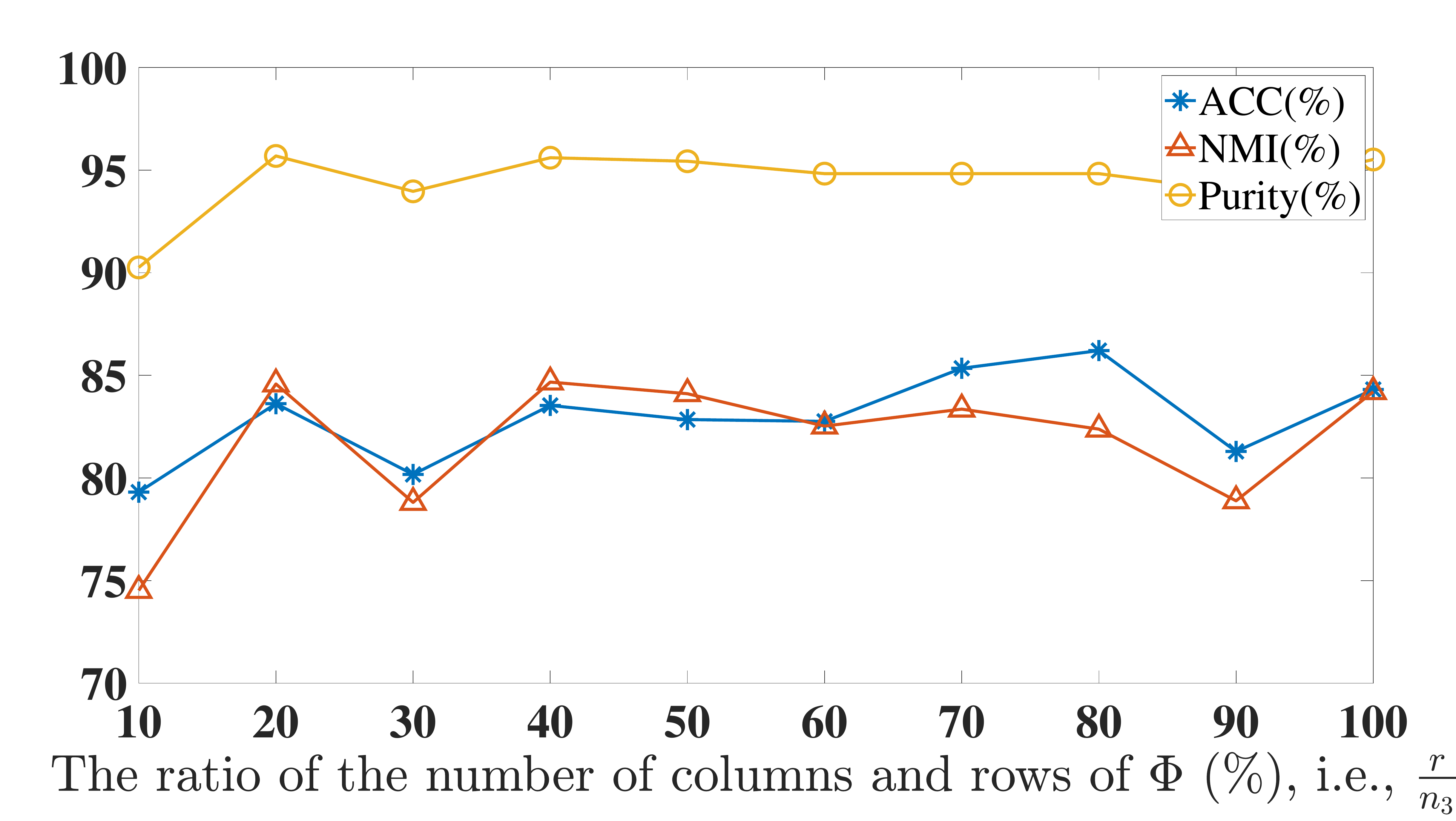}}
	\subfigure[]{
		\label{fignd} 
		\includegraphics[width=0.48\textwidth]{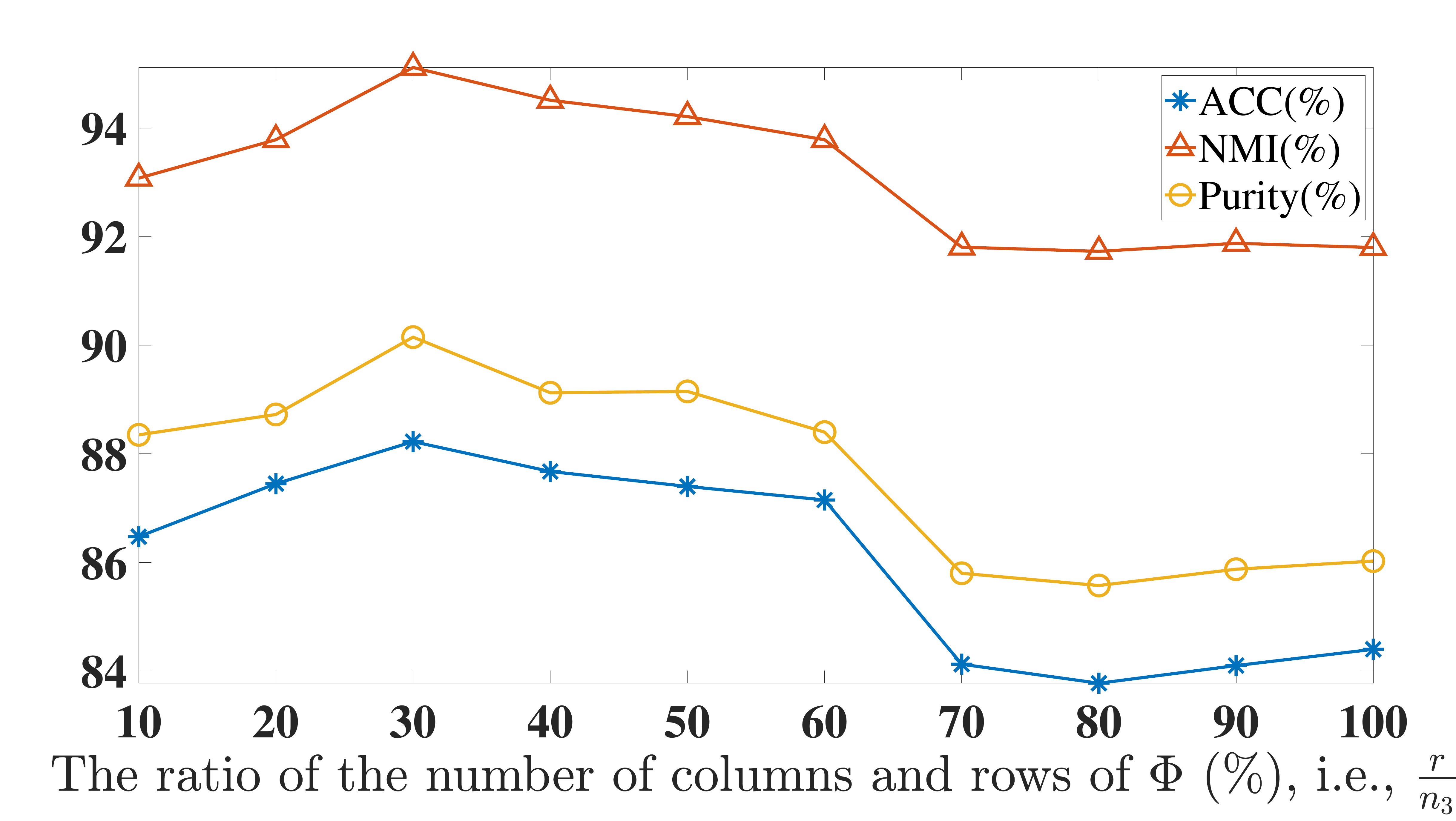}}
	\caption{The clustering performance for different proportions of parameter $r$ to $n_3$ on the (a) Handwritten, (b) 3 Sources, (c) BBC Sport, and (d) Orl datasets with a missing rate of $70\%$.}
	\label{fignormnn} 
\end{figure}

\begin{figure}[h]
	\centering
	\subfigure[]{
		\label{fignra} 
		\includegraphics[width=0.48\textwidth]{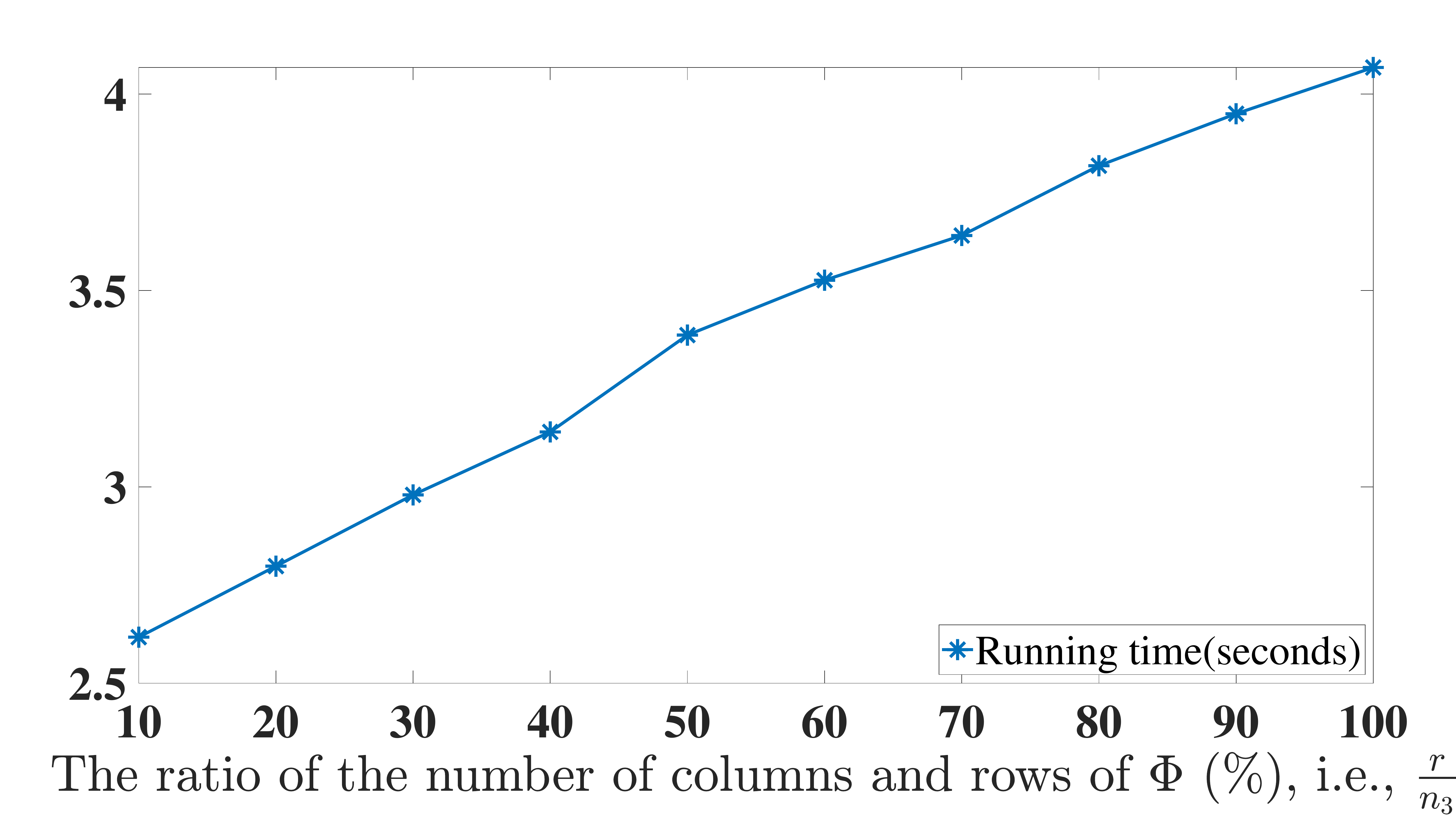}}
	\subfigure[]{
		\label{fignrb} 
		\includegraphics[width=0.48\textwidth]{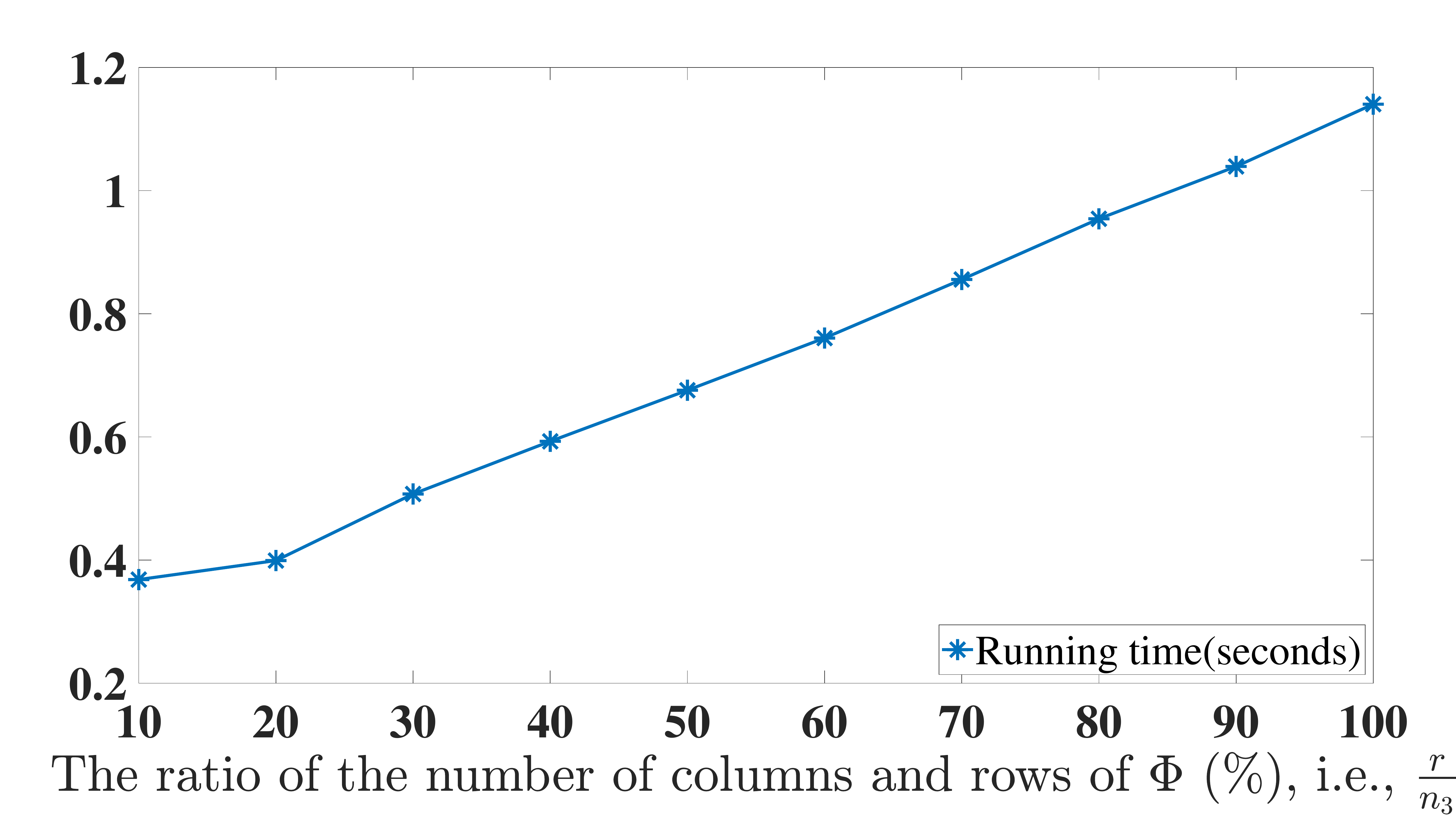}}\\
	\subfigure[]{
		\label{fignrc} 
		\includegraphics[width=0.48\textwidth]{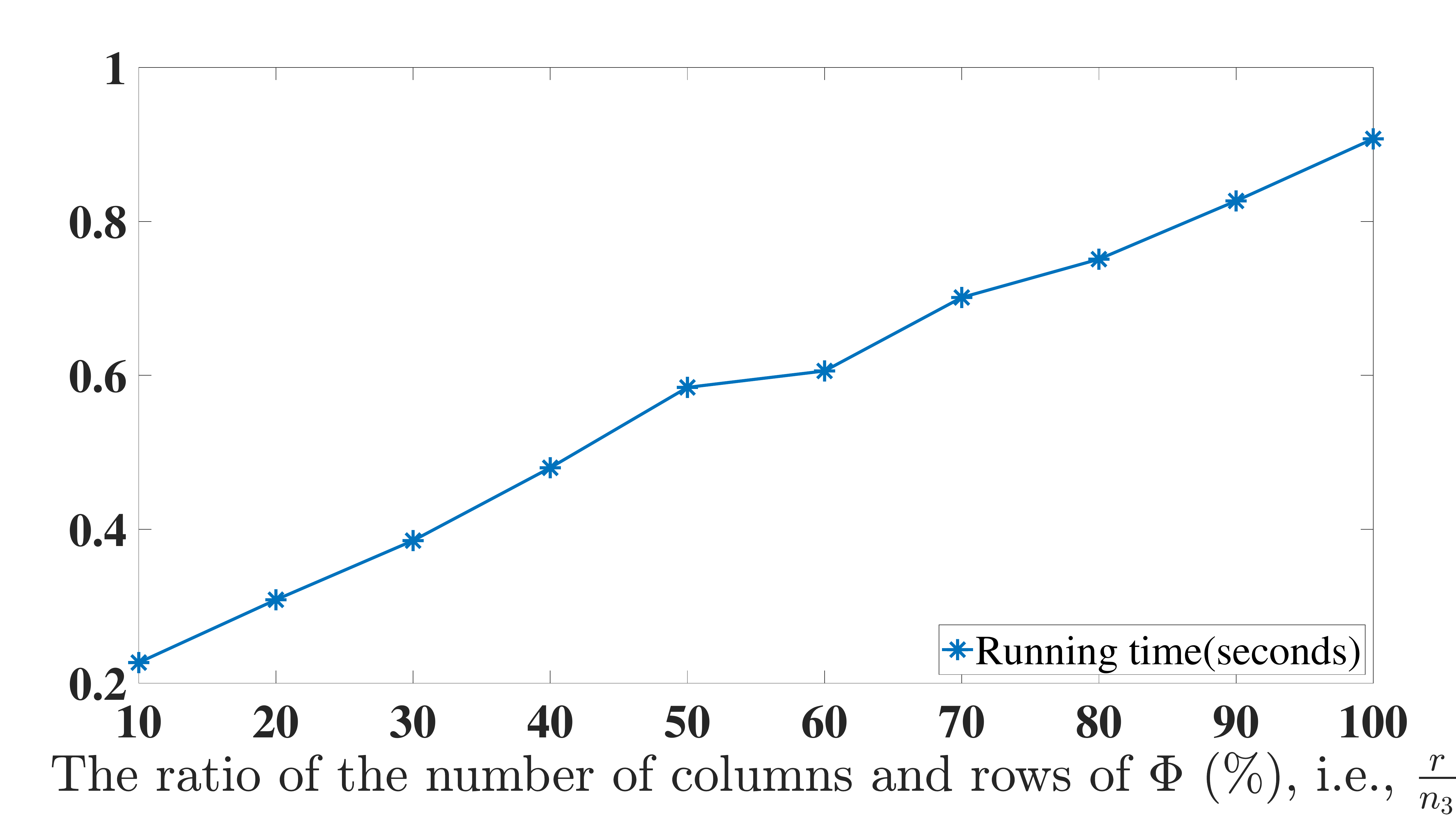}}
	\subfigure[]{
		\label{fignrd} 
		\includegraphics[width=0.48\textwidth]{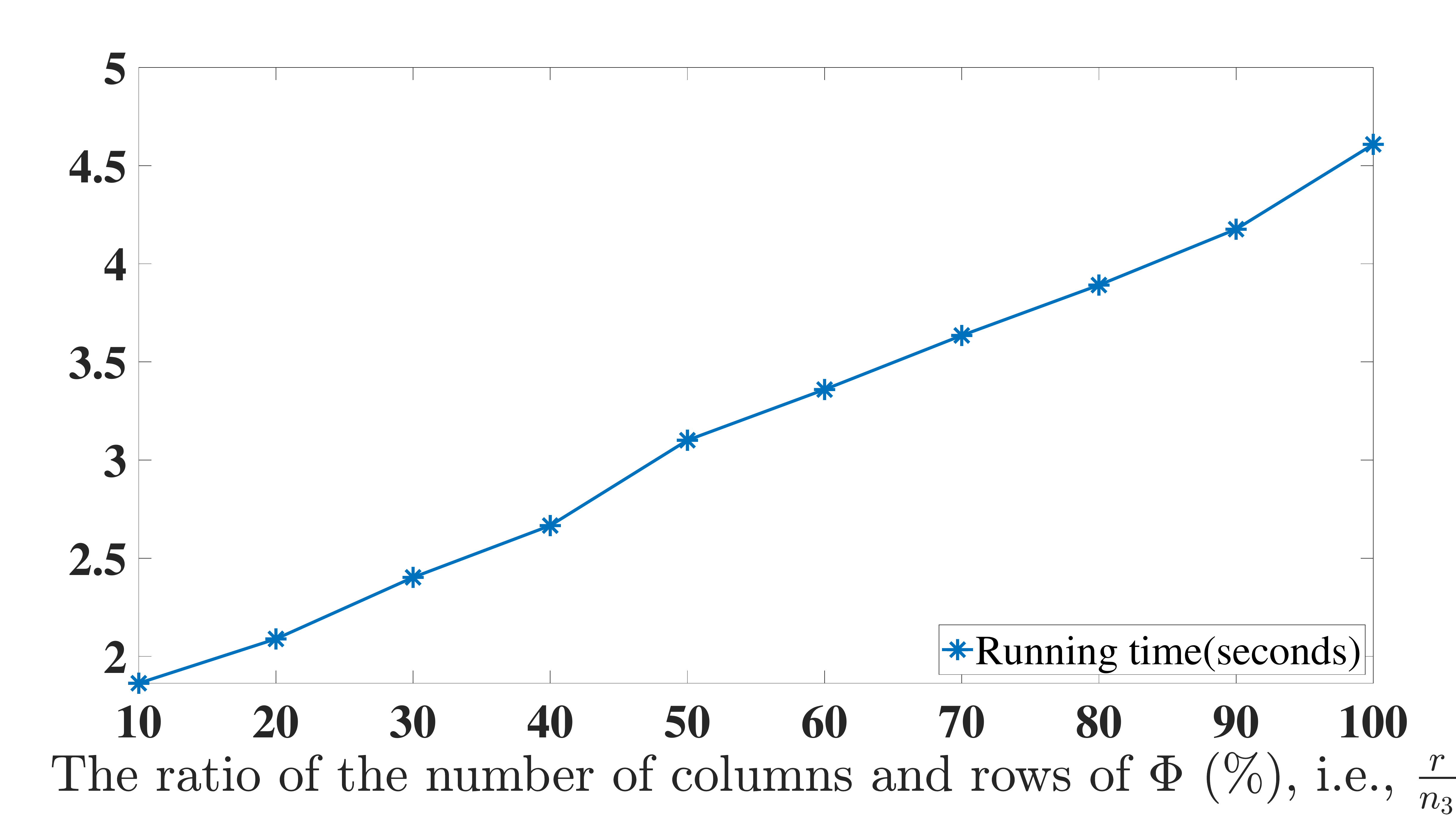}}
	\caption{The running time for different proportions of parameter $r$ to $n_3$ on the (a) Handwritten, (b) 3 Sources, (c) BBC Sport, and (d) Orl datasets with a missing rate of $70\%$.}
	\label{fignormruntime} 
\end{figure}

In this section, we investigate the effect of the parameter $r$ on the experimental result. We let $r$ be $10$ to $100$ percent of $n_3$ for each experiment on Handwritten, 3 Sources, BBC Sport, and Orl datasets with a missing rate of $70\%$, i.e., $r\in\left\{n_3\times10\%,n_3\times20\%,\dots,n_3\times100\%\right\}$. Figure \ref{fignormnn} shows that the clustering performance of our method on the Handwritten and Orl datasets tends to decrease with increasing $r/n_3$. The proposed model performance does not change significantly with increasing $r/n_3$ on the 3Sources and BBC Sport datasets. Figure \ref{fignormruntime} shows that the running time for solving $\mathcal{Y}$ becomes lager as $r/n_3$ increases on four datasets. In conclusion, $r$ taking the value of $n_3$ is not the optimal choice, so $r<n_3$ is meaningful in this paper.
\section{Conclusion}
\label{sec:conc}
In this paper, we present an incomplete multi-view learning method, which uses the completed graphs obtained by tensor recovery to guide the data project into consistent low-dimensional subspace to preserve the geometry among samples. At the same time, we provide an optimization method based on inexact augmented Lagrange multiplier (ALM) and analyze its convergence and computational complexity. Finally, we verify the effectiveness of our proposed method on real datasets.

\clearpage
\appendix
\section{Appendix}
\subsection{Proof of Theorem \ref{thm1}}
\label{A1}
Before proving Theorem \ref{thm1}, we first give Lemma \ref{lem1} and Lemma \ref{lem2}.

\begin{lemma}\cite{WPNMIDBS}
	\label{lem1}
	For the optimization problem
	\begin{equation}\label{equl12}
		\underset{x>0}{\min}\quad \frac{1}{2}\left(x-\sigma\right)^2+w x^p
	\end{equation}
	with the given $p$ and $w$, there is a specific threshold
	\begin{equation}
		\tau_p\left(w\right)=\left(2 w\left(1-p\right)\right)^{\frac{1}{2-p}}+w p\left(2 w\left(1-p\right)\right)^{\frac{p-1}{2-p}}
	\end{equation}
	we have the following conclusion
	\begin{enumerate}
		\item When $\sigma\le\tau_p\left(w\right)$, the optimal solution $x_p\left(\sigma, w\right)$ of Eq.(\ref{equl12}) is 0.
		\item When $\sigma>\tau_{p}\left(w\right)$, the optimal solution is $x_{p}\left(\sigma, w\right)=$ $\operatorname{sign}\left(\sigma\right) S_{p}\left(\sigma, w\right)$, where $S_{p}\left(\sigma, w\right)$ can be obtain by solving $S_{p}\left(\sigma, w\right)-\sigma+w p\left(S_{p}\left(\sigma, w\right)\right)^{p-1}=0 .$
	\end{enumerate}
\end{lemma}
\begin{lemma}\cite{WPNMIDBS}
	\label{lem2}
	Let $Y=U_{Y} \Sigma_{Y} V_{Y}^{\top}$ be the SVD of $Y\in \mathbb{R}^{m \times n}, \tau>0$, $l=\min (m, n), w_{1} \geq w_{2} \geq \cdots \geq w_{l}\geq0$, a global optimal solution of the following problem,
	\begin{equation}
		\underset{X}{\min } \quad \tau\|X\|_{w, S_{p}}^{p}+\frac{1}{2}\|X-Y\|_{F}^{2},
	\end{equation}
	is
	\begin{equation}
		S_{\tau,w,p}\left(Y\right)=U_{Y} P_{\tau,w,p}\left(Y\right) V_{Y}^{\top}
	\end{equation}
	where $P_{\tau,w,p}\left(Y\right)=\operatorname{diag}\left(\gamma_{1}, \gamma_{2}, \ldots,\gamma_{l}\right)$ and $\gamma_{i}=x_p\left(\sigma_i\left(Y\right), \tau w_{i}\right)$ which can be obtained by Lemma \ref{lem1}.
\end{lemma}

The proof of the Theorem \ref{thm1} is given as follows.

\begin{proof}
	For a given semi-orthogonal matrix $\Phi\in\mathbb{R}^{n_3\times r}$, there exists a semi-orthogonal matrix $\Phi^c\in\mathbb{R}^{n_3\times\left(n_3-r\right)}$ satisfying $\overline{\Phi}^\top\overline{\Phi}=I$, where $\overline{\Phi}=\left[\Phi,\Phi^c\right]\in\mathbb{R}^{n_3\times n_3}$.According to definition, we have
	\begin{equation}
		\label{pfthm11}
		\begin{split}
			\mathcal{X}^*=&\underset{\mathcal{X}}{\arg\min}\quad \tau\left\|\mathcal{X}\right\|_{\Phi, w, S_{p}}^{p}+\frac{1}{2}\left\|\mathcal{X}-\mathcal{A}\right\|_{F}^{2}\\
			=&\underset{\mathcal{X}}{\arg\min}\quad \tau\sum\limits_{i=1}^{r}\left\|X_{\Phi}^{\left(i\right)}\right\|_{w, S_{p}}^{p}+\frac{1}{2}\left\|\mathcal{X}_{\overline{\Phi}}-\mathcal{A}_{\overline{\Phi}}\right\|_{F}^{2}\\
			=&\underset{\mathcal{X}}{\arg\min}\quad \tau\sum\limits_{i=1}^{r}\left\|X_{\Phi}^{(i)}\right\|_{w, S_{p}}^{p}+\frac{1}{2}\sum\limits_{i=1}^r\left\|X_{\Phi}^{(i)}-A_{\Phi}^{(i)}\right\|_F^2+\frac{1}{2}\sum\limits_{j=1}^{n_3-r}\left\|X_{\Phi^c}^{(j)}-A_{\Phi^c}^{(j)}\right\|_F^2
		\end{split}		
	\end{equation}
	In Eq.(\ref{pfthm11}), each variable$X_{\Phi}^{(i)}$ and $X_{\Phi^c}^{(j)}$ is independent. Thus, it can be divided into $n_3$ independent subproblem, i.e.,
	\begin{equation}
		\label{pfthm12}
		\underset{\mathcal{X}}{\min}\quad \tau\left\|X_{\Phi}^{(i)}\right\|_{w, S_{p}}^{p}+\frac{1}{2}\left\|X_{\Phi}^{(i)}-A_{\Phi}^{(i)}\right\|_F^2, \qquad i=1,2,\cdots,r
	\end{equation}
	and
	\begin{equation}
		\label{pfthm13}
		\underset{\mathcal{X}}{\min}\quad \frac{1}{2}\left\|X_{\Phi^c}^{(j)}-A_{{\Phi^c}}^{(j)}\right\|_F^2,\qquad j=1,2,\cdots,n_3-r
	\end{equation}
	According to Lemma \ref{lem2}, the optimal solution of Eq.(\ref{pfthm12}) and Eq.(\ref{pfthm13}) are ${X_{\Phi}^{(i)}}^*=U_{A_{\Phi}^{(i)}} P_{\tau,w,p}\left(A_{\Phi}^{(i)}\right) V_{A_{\Phi}^{(i)}}^{\top}$ and ${X_{\Phi^c}^{(j)}}^*=A_{\Phi^c}^{(j)}$, respectively. Therefore, we obtain optimal
	\begin{equation}
		\begin{split}
			\mathcal{X}_{\overline{\Phi}}^*&=\operatorname{fold}\left({X_{\Phi}^{(1)}}^*; {X_{\Phi}^{(2)}}^*;\cdots;{X_{\Phi}^{(r)}}^*;{X_{\Phi^c}^{(1)}}^*;{X_{\Phi^c}^{(2)}}^*;\cdots;{X_{\Phi^c}^{(n_3-r)}}^*\right)\\
			&=\operatorname{fold}\left(S_{\tau,w,p}\left(A_{\Phi}^{(1)}\right);\cdots;S_{\tau,w,p}\left(A_{\Phi}^{(r)}\right);A_{\Phi^c}^{(1)};\cdots;A_{\Phi^c}^{(n_3-r)} \right)\\
			&=\operatorname{fold}\left(\operatorname{unfold}\left(\mathcal{S}_{\tau,w,p}\left(\mathcal{A}_{\Phi}\right)\right);\operatorname{unfold}\left(\mathcal{A}_{\Phi^c}\right)\right),
		\end{split}		
	\end{equation}
	then, further the optimal solution of Eq.(\ref{thm1equ}) is $\mathcal{X}^*=\left(\mathcal{X}_{\overline{\Phi}}^*\right)_{\overline{\Phi}^\top}$.
\end{proof}

\subsection{Proof of Lemma \ref{lemc}}
\label{A2}
\begin{proof}
	From the constraint on $\mathcal{G}$ in Eq.(\ref{our}), it is clear that $\left\{\mathcal{G}^k\right\}$ must be bounded. Further according to Eq.(\ref{updateA}) it is known that $\left\{A^k\right\}$ and $\left\{\left\{W_i^k\right\}_{i=1}^m\right\}$ are also bounded. In the $k$-th iteration in Algorithm \ref{alg}, we have
	\begin{equation}
		\label{prflem21}
		\begin{split}
			\left\|\mathcal{C}^{k+1}\right\|_F&=\left\|\mathcal{C}^{k}+\rho^k\left(\mathcal{G}^{k+1}-\mathcal{Y}^{k+1}\right)\right\|_F\\
			&=\rho^k\left\|\mathcal{G}^{k+1}+\frac{\mathcal{C}^k}{\rho^k}-\mathcal{Y}^{k+1}\right\|_F\\
			&=\rho^k\left\|\left(\mathcal{G}^{k+1}+\frac{\mathcal{C}^k}{\rho^k}-\mathcal{Y}^{k+1}\right)_{\overline{\Phi}}\right\|_F\\
			&=\rho^k\left\|\mathcal{B}_{\overline{\Phi}}^k-\mathcal{Y}^{k+1}_{\overline{\Phi}}\right\|_F.
		\end{split}
	\end{equation}
	From Eq.(\ref{updateY2}) we can get 
	\begin{equation}
		\label{prflem22}
		\mathcal{Y}_{\overline{\Phi}}^{k+1}=\operatorname{fold}\left(\operatorname{unfold}\left(\mathcal{S}_{\frac{\mu}{\rho^k},w,p}(\mathcal{B}_\Phi^{k})\right);\operatorname{unfold}\left(\mathcal{B}_{\Phi^c}^{k}\right)\right).
	\end{equation}
	Substituting Eq.(\ref{prflem22}) into Wq.(\ref{prflem21}), we have 
	\begin{equation}
		\begin{split}
			\left\|\mathcal{C}^{k+1}\right\|_F&=\rho^k\left\|\mathcal{B}_{\overline{\Phi}}^k-\operatorname{fold}\left(\operatorname{unfold}\left(\mathcal{S}_{\frac{\mu}{\rho^k},w,p}\left(\mathcal{B}_\Phi^{k}\right)\right);\operatorname{unfold}\left(\mathcal{B}_{\Phi^c}^{k}\right)\right)\right\|_F\\
			&=\rho^k\left\|\mathcal{B}_{\Phi}^k-\mathcal{S}_{\frac{\mu}{\rho^k},w,p}\left(\mathcal{B}_\Phi^{k}\right)\right\|_F\\
			&\leq\rho^k\sqrt{\sum\limits_{i=1}^{r}\sum\limits_{j=1}^{min\left(n_1,n_2\right)}\left(\frac{Jw_j \mu}{\rho^k}\right)^2}\\
			&=\sqrt{\sum\limits_{i=1}^{r}\sum\limits_{j=1}^{min\left(n_1,n_2\right)}\left(Jw_j \mu\right)^2},
		\end{split}
	\end{equation}
	where $J$ is the number of iterations in solving the problem (\ref{equl12}). Thus, $\left\{\mathcal{C}^k\right\}$ is bounded. Further according to Eq.(\ref{updateY2}) we can see that 	
	\begin{equation}
		\begin{split}
			\left\|\mathcal{Y}^{k+1}\right\|_F&=\left\|\operatorname{fold}\left(\operatorname{unfold}\left(\mathcal{S}_{\frac{\mu}{\rho},w,p}(\mathcal{B}^{k}_\Phi)\right);\operatorname{unfold}\left(\mathcal{B}^{k}_{\Phi^c}\right)\right)\right\|_F\\
			&\leq\left\|\mathcal{S}_{\frac{\mu}{\rho},w,p}\left(\mathcal{B}^{k}_\Phi\right)\right\|_F+\left\|\mathcal{B}^{k}_{\Phi^c}\right\|_F\\
			&\leq\sum\limits_{i=1}^r\left\|P_{\tau,w,p}\left({B_\Phi^k}^{\left(i\right)}\right)\right\|_F+\left\|\mathcal{B}^{k}_{\Phi^c}\right\|_F.
		\end{split}
	\end{equation}
	It has already been mentioned that both $\left\{\mathcal{C}^k\right\}$ and $\left\{\mathcal{G}^k\right\}$ are bounded, so $\left\{\mathcal{B}^k\right\}$ is bounded, and thus $\left\{\mathcal{Y}^k\right\}$ is bounded.	
	
	The following proves that the augmented Lagrangian function Eq.(\ref{largfun}) is bounded\footnote{Note that one cannot state that Eq.(\ref{largfun}) is bounded based on the fact that all variables are bounded due to $\left\{\rho^k\right\}$ is unbounded}. Let's first to prove 
	\begin{equation}
		\label{prflem20}
		L_{\rho^k}\left(A^{k+1}, \left\{W_i^{k+1}\right\}_{i=1}^m, \mathcal{G}^k, \mathcal{Y}^k, \mathcal{C}^k\right)\leq L_{\rho^k}\left(A^k, \left\{W_i^k\right\}_{i=1}^m, \mathcal{G}^k, \mathcal{Y}^k, \mathcal{C}^k\right).
	\end{equation}
	
	Recall Eq.(\ref{updataAW}), let $f\left(A, W_i\right)=\left\|\left(A-{W_i}^\top X_i\right)P_i\right\|_F^2+\lambda \operatorname{tr}\left(AL_{G_i^k}A^\top\right)$ and combining with $\delta_i^k=n_i/\sqrt{f\left(A^k, W_i^k\right)}$, we can derive
	\begin{equation}
		\label{prflem23}
		\sum\limits_{i=1}^m\frac{n_if\left(A^{k+1}, W_i^{k+1}\right)}{2\sqrt{f\left(A^k, W_i^k\right)}}\leq \sum\limits_{i=1}^m\frac{n_if\left(A^k, W_i^k\right)}{2\sqrt{f\left(A^k, W_i^k\right)}}.
	\end{equation}
	And since $f\left(A, W_i\right)\geq 0$, we have\footnote{\begin{corollary}
			For any positive number $a$ and $b$, the following inequality holds
			\begin{equation}
				a-\frac{a^2}{2b}\leq b-\frac{b^2}{2b}.
			\end{equation}
	\end{corollary}}
	\begin{equation}
		\label{prflem24}
		\sum\limits_{i=1}^mn_i\left(\sqrt{f\left(A^{k+1},W_i^{k+1}\right)}-\frac{f\left(A^{k+1}, W_i^{k+1}\right)}{2\sqrt{f\left(A^k, W_i^k\right)}}\right)\leq \sum\limits_{i=1}^mn_i\left(\sqrt{f\left(A^k,W_i^k\right)}- \frac{f\left(A^k, W_i^k\right)}{2\sqrt{f\left(A^k, W_i^k\right)}}\right).
	\end{equation}
	Combining Eq.(\ref{prflem23}) and Eq.(\ref{prflem24}), we arrive at 
	\begin{equation}
		\sum\limits_{i=1}^mn_i\sqrt{f\left(A^{k+1},W_i^{k+1}\right)}\leq \sum\limits_{i=1}^mn_i\sqrt{f\left(A^k,W_i^k\right)}.
	\end{equation}
	Therefore, Eq.(\ref{prflem20}) is derived. In section \ref{sec:updateG} and section \ref{sec:updateY}, since both $\mathcal{G}$ and $\mathcal{Y}$ subproblem have optimal solution, we have obviously
	\begin{equation}
		\label{prflem25}
		\begin{split}
			L_{\rho^k}\left(A^{k+1}, \left\{W_i^{k+1}\right\}_{i=1}^m, \mathcal{G}^{k+1}, \mathcal{Y}^{k+1}, \mathcal{C}^k\right)&\leq L_{\rho^k}\left(A^{k+1}, \left\{W_i^{k+1}\right\}_{i=1}^m, \mathcal{G}^k, \mathcal{Y}^k, \mathcal{C}^k\right)\\
			&\leq L_{\rho^k}\left(A^k, \left\{W_i^k\right\}_{i=1}^m, \mathcal{G}^k, \mathcal{Y}^k, \mathcal{C}^k\right).
		\end{split}
	\end{equation}
	Then 
	\begin{equation}
		\label{prflem26}
		\begin{split}
			L_{\rho^k}\left(A^k, \left\{W_i^k\right\}_{i=1}^m, \mathcal{G}^k, \mathcal{Y}^k, \mathcal{C}^k\right)&=L_{\rho^{k-1}}\left(A^k, \left\{W_i^k\right\}_{i=1}^m, \mathcal{G}^k, \mathcal{Y}^k,\mathcal{C}^{k-1}\right)+\frac{\rho^k-\rho^{k-1}}{2}\left\|\mathcal{G}^k-\mathcal{Y}^k\right\|_F^2\\
			&\quad+\left\langle\mathcal{C}_1^k-\mathcal{C}_1^{k-1},\mathcal{G}^k-\mathcal{Y}^k\right\rangle\\
			&=L_{\rho^{k-1}}\left(A^k, \left\{W_i^k\right\}_{i=1}^m, \mathcal{G}^k, \mathcal{Y}^k, \mathcal{C}^{k-1}\right)\\
			&\quad+\frac{\rho^k-\rho^{k-1}}{2}\left\|\frac{\mathcal{C}^k-\mathcal{C}^{k-1}}{\rho^{k-1}}\right\|_F^2+\left\langle\mathcal{C}^k-\mathcal{C}^{k-1},\frac{\mathcal{C}^k-\mathcal{C}^{k-1}}{\rho^{k-1}}\right\rangle\\
			&=L_{\rho^{k-1}}\left(A^k, \left\{W_i^k\right\}_{i=1}^m, \mathcal{G}^k, \mathcal{Y}^k, \mathcal{C}^{k-1}\right)+\frac{\rho^k+\rho^{k-1}}{2{\rho^{k-1}}^2}\left\|\mathcal{C}^k-\mathcal{C}^{k-1}\right\|_F^2.
		\end{split}
	\end{equation}
	Denote by $b_c$ the bound of $\left\|\mathcal{C}^k-\mathcal{C}^{k-1}\right\|_F^2$, and combining Eq.(\ref{prflem25}) we have 
	\begin{equation}
		\label{prflem27}
		L_{\rho^k}\left(A^{k+1}, \left\{W_i^{k+1}\right\}_{i=1}^m, \mathcal{G}^{k+1}, \mathcal{Y}^{k+1}, \mathcal{C}^k\right)\leq L_{\rho^0}\left(A^1, \left\{W_i^1\right\}_{i=1}^m, \mathcal{G}^1, \mathcal{Y}^1, \mathcal{C}^0, \right)+b_c\sum\limits_{k=1}^\infty \frac{\rho^k+\rho^{k-1}}{2{\rho^{k-1}}^2}.
	\end{equation}
	Since $\rho>1$, the following inequality holds,
	\begin{equation}
		\label{prflem28}
		\sum\limits_{k=1}^\infty \frac{\rho^k+\rho^{k-1}}{2{\rho^{k-1}}^2}\leq \sum\limits_{k=1}^\infty\frac{\rho^k}{{\rho^{k-1}}^2}=\alpha\sum\limits_{k=1}^\infty\frac{1}{\rho^{k-1}}<+\infty.
	\end{equation}
	Combining Eq.(\ref{prflem26}), Eq.(\ref{prflem27}), and Eq.(\ref{prflem28}), it is known that the augmented Lagrangian function Eq.(\ref{largfun}) is bounded.
\end{proof}

\subsection{Proof of Theorem \ref{thm2}}
\label{A3}
\begin{proof}
	According to Lemma \ref{lemc}, we know $\left\{\mathcal{C}^k\right\}$, $\left\{\mathcal{G}^k\right\}$, and $\left\{\mathcal{Y}^k\right\}$ are all bounded. Bolzano-Weierstrass Theorem \cite{TEBWT} shows that every bounded sequence of real numbers has a convergent subsequence. So there exists at least one accumulation point for $\left\{\mathcal{C}^k\right\}$, $\left\{\mathcal{G}^k\right\}$, and $\left\{\mathcal{Y}^k\right\}$. Specifically, we can get
	\begin{equation}
		\label{limGY}
		\lim\limits_{k\to\infty}\left\|\mathcal{G}^k-\mathcal{Y}^k\right\|_F=\lim\limits_{k\to\infty}\frac{1}{\rho^k}\left\|\mathcal{C}_1^k-\mathcal{C}_1^{k-1}\right\|_F=0.
	\end{equation}
	Similar to section \ref{sec:updateG}, here again for each tube of $\mathcal{G}^k$ is analyzed independently. Recalling that $g^k$ is updated in the way Eq.(\ref{updategk}), the following inequality holds,
	\begin{equation}
		\label{prfthm21}
		\begin{split}
			\left\|g^{k+1}-g^k\right\|_F&=\left\|\begin{split}
				&\frac{\rho^k}{\rho^k+\gamma}\operatorname{max}\left(P_w\left(u^k-\frac{1}{n}\textbf{1}\textbf{1}^\top u^k+\frac{1}{n}\textbf{1}+v\textbf{1}\right),0\right)\\
				&+\operatorname{max}\left(P_{w^c}\left(u^k-\frac{1}{n}\textbf{1}\textbf{1}^\top u^k+\frac{1}{n}\textbf{1}+v\textbf{1}\right),0\right)-g^k
			\end{split}\right\|_F\\
			&\leq\left\|\frac{\rho^k}{\rho^k+\gamma}\left(u^k-\frac{1}{n}\textbf{1}\textbf{1}^\top u^k+\frac{1}{n}\textbf{1}+v\textbf{1}-\frac{\rho^k+\gamma}{\rho^k}g^k\right)\right\|_F\\
			&\quad+\left\|u^k-\frac{1}{n}\textbf{1}\textbf{1}^\top u^k+\frac{1}{n}\textbf{1}+v\textbf{1}-g^k\right\|_F.
		\end{split}
	\end{equation}
	We analyze the first term and second term after the inequality sign of the Eq.(\ref{prfthm21}) separately as follows,

	\begin{equation}
		\begin{split}
			&\left\|\frac{\rho^k}{\rho^k+\gamma}\left(u^k-\frac{1}{n}\textbf{1}\textbf{1}^\top u^k+\frac{1}{n}\textbf{1}+v\textbf{1}-\frac{\rho^k+\gamma}{\rho^k}g^k\right)\right\|_F\\
			=&\left\|\frac{\rho^k}{\rho^k+\gamma}\left(y^k-\frac{c^k_1}{\rho^k}+\frac{\gamma}{\rho^k}P_w\left(m\right)-\frac{1}{\rho^k}t^k-\frac{1}{n}\textbf{1}\textbf{1}^\top\left(y^k-\frac{c^k_1}{\rho^k}+\frac{\gamma}{\rho^k}P_w\left(m\right)-\frac{1}{\rho^k}t^k\right)+\frac{1}{n}\textbf{1}-g^k-\frac{\gamma}{\rho^k}g^k+v\textbf{1}\right)\right\|_F\\
			=&\left\|\frac{\rho^k}{\rho^k+\gamma}\left(\frac{c_1^{k-1}-2c_1^k}{\rho^k}+\frac{\gamma}{\rho^k}P_w\left(m\right)-\frac{1}{\rho^k}t^k-\frac{1}{n}\textbf{1}\textbf{1}^\top\left(\frac{c^{k-1}_1-2c_1^k}{\rho^k}+\frac{\gamma}{\rho^k}P_w\left(m\right)-\frac{1}{\rho^k}t^k\right)-\frac{\gamma}{\rho^k}g^k+v\textbf{1}\right)\right\|_F,
		\end{split}
	\end{equation}
	and
	\begin{equation}
		\begin{split}
			&\left\|u^k-\frac{1}{n}\textbf{1}\textbf{1}^\top u^k+\frac{1}{n}\textbf{1}+v\textbf{1}-g^k\right\|_F\\
			=&\left\|y^k-\frac{c^k_1}{\rho^k}+\frac{\gamma}{\rho^k}P_w\left(m\right)-\frac{1}{\rho^k}t^k-\frac{1}{n}\textbf{1}\textbf{1}^\top \left(y^k-\frac{c^k_1}{\rho^k}+\frac{\gamma}{\rho^k}P_w\left(m\right)-\frac{1}{\rho^k}t^k\right)+\frac{1}{n}\textbf{1}+v\textbf{1}-g^k\right\|_F\\
			=&\left\|\frac{c^{k-1}_1-c^k_1}{\rho^k}+\frac{\gamma}{\rho^k}P_w\left(m\right)-\frac{1}{\rho^k}t^k-\frac{1}{n}\textbf{1}\textbf{1}^\top \left(\frac{c^{k-1}_1-c^k_1}{\rho^k}+\frac{\gamma}{\rho^k}P_w\left(m\right)-\frac{1}{\rho^k}t^k\right)+v\textbf{1}\right\|_F.
		\end{split}
	\end{equation}
	The following shows that $v$ tends to $0$ when $k$ tends to infinity. According to the definition of $u^k$ the following equation holds,
	\begin{equation}
		\label{limitiug}
		\lim\limits_{k \to \infty}\left\|u^k-g^k\right\|_2=\left\|g^k+\frac{c_1^{k-1}-2c_1^k}{\rho^k}+\frac{\gamma}{\rho^k}P_w\left(m\right)-\frac{1}{\rho^k}t^k-g^k\right\|_2=0.
	\end{equation}
	Combining Eq.(\ref{updategk}) and Eq.(\ref{sov}), Eq.(\ref{limitiug}) implies that $f\left(0\right)=0$ when $k$ tends to infinity, i.e. $v$ tends to $0$.
	In summary, $\lim\limits_{k\to\infty}\left\|g^{k+1}-g^k\right\|_2=0$ and hence
	\begin{equation}
		\label{limGG}
		\lim\limits_{k\to\infty}\left\|\mathcal{G}^{k+1}-\mathcal{G}^k\right\|_F=0.
	\end{equation}
	The following inequality holds for $\mathcal{Y}^k$,
	\begin{equation}
		\begin{split}
			\left\|\mathcal{Y}^{k+1}-\mathcal{Y}^k\right\|_F&=\left\|\mathcal{Y}^{k+1}-\mathcal{G}^{k+1}-\mathcal{Y}^k+\mathcal{G}^k+\mathcal{G}^{k+1}-\mathcal{G}^k\right\|_F\\
			&\leq\left\|\mathcal{Y}^{k+1}-\mathcal{G}^{k+1}\right\|+\left\|\mathcal{Y}^k-\mathcal{G}^k\right\|+\left\|\mathcal{G}^{k+1}-\mathcal{G}^k\right\|_F.
		\end{split}
	\end{equation}
	Combining Eq.(\ref{limGY}) and Eq.(\ref{limGG}), therefore, we have
	\begin{equation}
		\lim\limits_{k\to\infty}\left\|\mathcal{Y}^{k+1}-\mathcal{Y}^k\right\|_F=0.
	\end{equation}
\end{proof}

\section*{Acknowledgments}
We would like to acknowledge the assistance of volunteers in putting
together this example manuscript and supplement.

\bibliographystyle{unsrt}

\bibliography{Bibliography}

\end{document}